\documentclass[journal]{IEEEtran}

\usepackage{tikz}

\newcommand\copyrighttext{%
  \footnotesize \textcopyright 2024 IEEE. Personal use of this material is permitted.
  Permission from IEEE must be obtained for all other uses, in any current or future 
  media, including reprinting/republishing this material for advertising or promotional 
  purposes, creating new collective works, for resale or redistribution to servers or 
  lists, or reuse of any copyrighted component of this work in other works.  
  }
\newcommand\copyrightnotice{%
\begin{tikzpicture}[remember picture,overlay]
\node[anchor=south,yshift=10pt] at (current page.south) {\fbox{\parbox{\dimexpr\textwidth-\fboxsep-\fboxrule\relax}{\copyrighttext}}};
\end{tikzpicture}%
}

\usepackage{amsmath,amsfonts}
\usepackage{algorithmic}
\usepackage{array}
\ifCLASSOPTIONcompsoc
    \usepackage[caption=false, font=normalsize, labelfont=sf, textfont=sf]{subfig}
\else
\usepackage[caption=false, font=footnotesize]{subfig}
\fi
\usepackage{textcomp}
\usepackage{stfloats}
\usepackage{url}
\usepackage{verbatim}
\usepackage{graphicx}
\usepackage{balance}
\usepackage{hyperref}
\usepackage{gensymb}

\usepackage{xcolor}

\newcommand \W{\mathcal{W}}
\newcommand \OM{\mathcal{\tilde{W}}}
\newcommand \E{\mathcal{E}}

\newcommand \C{\mathnormal{C}}

\makeatletter

\def\BibTeX{{\rm B\kern-.05em{\sc i\kern-.025em b}\kern-.08em T\kern-.1667em\lower.7ex\hbox{E}\kern-.125emX}}
\begin{document}
\title{PEM: Perception Error Model for Virtual Testing of Autonomous Vehicles}

\author{Andrea Piazzoni, Jim Cherian, Justin Dauwels,
\IEEEmembership{Senior Member, IEEE},
Lap-Pui Chau,
\IEEEmembership{Fellow, IEEE}
 \thanks{
This work was supported by the National Research Foundation, Singapore, and Land Transport Authority through Urban Mobility Grand Challenge under Grant UMGC-L010.}
  \thanks{Andrea Piazzoni is with the Interdisciplinary Graduate Programme - ERI@N, 
 Centre of Excellence for Testing \& Research of AVs, Nanyang Technological University, Singapore (e-mail: apiazzoni@ntu.edu.sg).}
 \thanks{Jim Cherian is with the Advanced Remanufacturing \& Technology Centre (ARTC)
Agency for Science, Technology and Research (A*STAR), Singapore
 (e-mail: jimcherian.wk@gmail.com).}
 \thanks{Justin Dauwels is with TU Delft, Department of Microelectronics, Fac. EEMCS, Mekelweg 4 2628 CD, Delft. (e-mail: j.h.g.dauwels@tudelft.nl).}
  \thanks{Lap-Pui Chau is with the Department of Electronic and Information Engineering, The Hong Kong Polytechnic University (e-mail: elpchau@ntu.edu.sg).}
  }

\maketitle

\copyrightnotice

\begin{abstract}
Even though virtual testing of Autonomous Vehicles (AVs) has been well recognized as essential for safety assessment, AV simulators are still undergoing active development.
One particularly challenging question is to effectively include the Sensing and Perception (S\&P) subsystem into the simulation loop.
In this article, we define Perception Error Models (PEM), a virtual simulation component that can enable the analysis of the impact of perception errors on AV safety, without the need to model the sensors themselves. We propose a generalized data-driven procedure towards parametric modeling and evaluate it using Apollo, an open-source driving software, and nuScenes, a public AV dataset. Additionally, we implement PEMs in SVL, an open-source vehicle simulator.  Furthermore, we demonstrate the usefulness of PEM-based virtual tests, by evaluating camera, LiDAR, and camera-LiDAR setups. Our virtual tests highlight limitations in the current evaluation metrics, and the proposed approach can help study the impact of perception errors on AV safety.
 \end{abstract}
 \begin{IEEEkeywords} Autonomous vehicles, Computer vision, Vehicle safety, Simulation.
 \end{IEEEkeywords}

\section{Introduction}
\IEEEPARstart{O}{ne} of the major limitations preventing a public deployment of Autonomous Vehicles (AVs) is their safety.
Furthermore, Shen et al. \cite{shen2020explain} emphasize how the explainability of Artificial Intelligence (AI) in AVs is necessary for trust and acceptance.
Hence, the automotive industry, the academic community, and regulators are developing safety assessment procedures.
By analyzing the AV Operational Design Domain (ODD)~\cite{standard2018j3016}, we can identify the challenges that the AV will encounter. 
In particular, sensors are known to be very susceptible to weather conditions and the amount of sunlight~\cite{Marti2019}.
The AV should handle such issues and still exhibit safe driving behavior. 
For example, if the perception uncertainty increases, the AV could reduce its speed and adopt a more defensive driving, thus maintaining an adequate level of safety.
Nevertheless, failures in obstacle detection may still lead to undesirable behavior such as collisions, emergency maneuvers, or traffic rules violations. 
For instance, the leading cause of a 2018 AV fatal accident was determined to be a perception error that was not adequately handled \cite{NTSB2019}.
Thus, a deeper understanding of how perception errors affect the AV response is necessary for safety assurance.

This connection between perception errors and AV response can be explored via a holistic testing approach, both on a test track and in virtual environments. In this paper, we concentrate on virtual tests. 
Virtual testing of AVs by simulations offers a safe and convenient way to validate safety \cite{Young2014}. However, how to effectively include perception modules in the simulation pipeline is an open question.
A common approach in the industry is to employ high-fidelity models that represent the real world.  In particular, physics-based sensor simulations may generate synthetic sensor signals, which the perception module can process. 
Besides being highly compute-intensive, this approach is not yet sufficiently developed since it is challenging to simulate sensors at high physical fidelity.
Therefore, we aim to develop a less complex alternative by modeling the functionality of the S\&P in terms of perception errors, which we can conveniently integrate into a simulation pipeline.
\begin{figure*}[t]
     \centering
     \includegraphics[width=\textwidth]{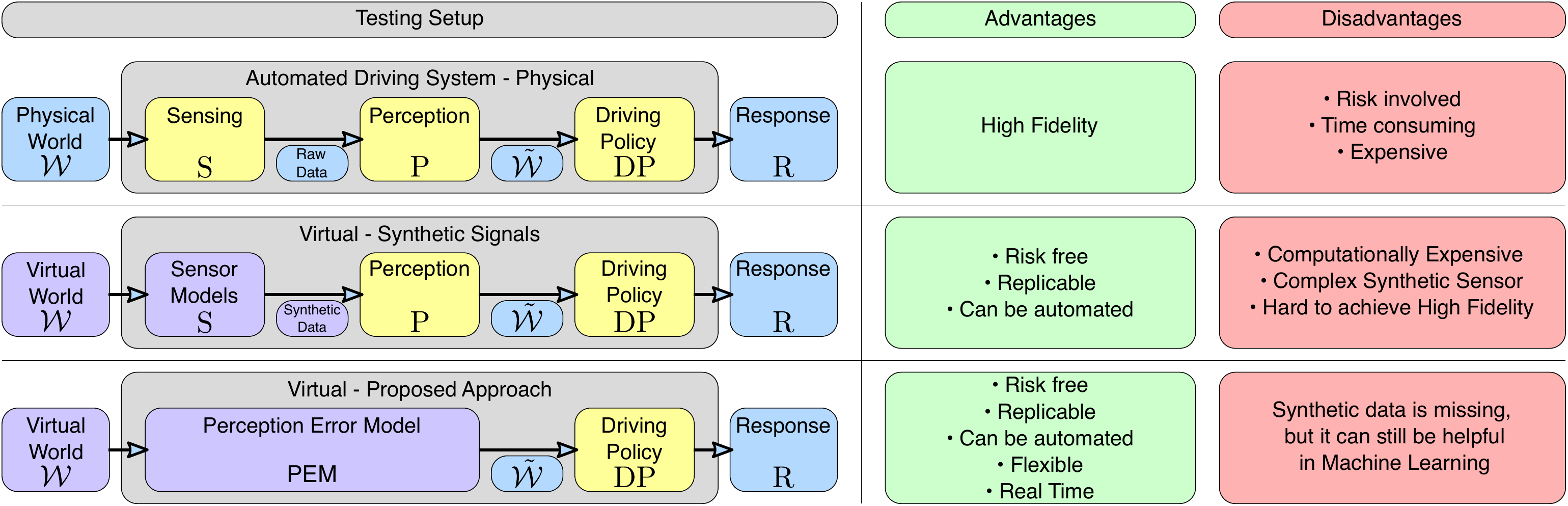}
     \caption{Comparison of three different testing setups that integrate perception: physical testing (top), synthetic signals in a virtual environment (middle), and our proposed approach involving PEMs (bottom), models that approximate the error of a Sensing and Perception module.}
     \label{fig:AV_pipeline}
\end{figure*}
In this article, we provide the following contributions:
\begin{itemize}
    \item We summarize the state-of-the-art evaluation metrics for AI-based perception algorithms and discuss how they are inadequate for assessing the safety of AVs. 
    \item We present a simulation pipeline that integrates PEMs in a virtual environment as a computationally efficient way of testing \mbox{Automated Driving System} (ADS) capabilities w.r.t. perception errors. 
    \item We present a generalized data-driven approach towards building Perception Error Models (PEMs), using public datasets and Open-Source software.
    \item We demonstrate the application of PEMs by comparing three different sensor setups, i.e., camera, LiDAR, and fusion of camera and LiDAR. Moreover, we assess the safety provided by these setups in three representative urban scenarios.
\end{itemize}

Our previous work relied on hand-crafted PEMs, which helped demonstrate a few limitations of evaluation metrics \cite{ijcai2020-483}. This article extends that work by introducing a data-driven modeling step and generalizing the procedure. This way, PEMs deployed in simulation experiments provide meaningful results. Thus, we present a complete procedure to investigate the impact of perception errors on safety. 

This article is organized as follows. In Sections 2 and 3, respectively, we provide an overview of the literature on Virtual Testing and Perception Evaluation and Modeling. in Section 4, we introduce our proposed definition of PEM, while in Section 5, we illustrate this definition with an example of a PEM. In Section 6, we demonstrate how a PEM can be integrated into a simulation pipeline. At last, in Section 7, we discuss our study, and in Section 8, we summarize the contributions and future directions. In \autoref{tab:Acronym} we report all the acronyms we adopt throughout the article.

\begin{table}[t]
    \centering
        \caption{Acronyms}
    \label{tab:Acronym}
    \begin{tabular}{c|c}
    Acronym &Meaning \\
    \hline
    ADS & Automated Driving System\\
    AV & Autonomous Vehicle\\
      CV & Computer Vision\\
        DP & Driving Policy\\
        FoV & Field of View\\
        IoU&Intersection Over Union\\
        mAP&Mean Average Precision\\
        MLE& Maximum Likelihood Estimation\\
        ODD & Operational Design Domain\\
        PEM& Percepion Error Model\\
      S\&P & Sensing and Perception\\
        $\W$& World, i.e. surrounding obstacles\\
        $\OM$& Perceived World\\
    \end{tabular}
\end{table}
\section{Basic Concepts of Virtual Testing}
\noindent In this Section, we describe our abstraction of an AV. Then, we provide an overview of AV virtual simulations and testing. 

\subsection{AV Modules}
As illustrated in \autoref{fig:AV_pipeline}, we describe AVs through three components: Sensing (S), Perception (P), and Driving Policy (DP).
The Sensing component $S$, or sensor setups, comprises all the AV's sensors to sense the surroundings. Standard sensors in the AV field are cameras, LiDAR, and RADAR \cite{rosique2019systematic}. 
The Perception component $P$ algorithm processes sensor signals to provide the perceived world $\OM$ (or object list), which is usually a noisy version of the ground truth world $\W$.
With the term Driving Policy DP \cite{Shalev2017formal}, we refer to the process that determines the response R of the AV given the perceived state of the physical world, i.e., the AV behavior.

\subsection{Virtual Environments and Simulations} \label{ssec:virtualenv}
 Virtual simulations are useful for various reasons (see \autoref{fig:AV_pipeline}). First, they provide a risk-free environment, which serves to avoid fatalities and damages to properties. 
 Moreover, numerous simulations can be run, and thousands of scenarios can be tested, very often with high repeatability. Therefore, virtual testing is a practical and economical approach for developers and third-party evaluators alike.
 
Key ingredients for virtual testing are a set of test scenarios that replicate real-life situations that AVs may be exposed to \cite{feng2020testing,piazzoni2021vista,gomez2022train}, and measurable safety metrics \cite{guo2019safe}.
To assess AVs by virtual testing, evaluators compute safety metrics from the simulated AV behaviors, such as safety clearance distances (between AV and other traffic participants), speed limits, or specific traffic rules violations.
Thresholds on those safety metrics demark safe from unsafe behavior. 
The major drawback of virtual testing is its potential lack of fidelity. 
Key constraints here include the modeling effort and computational costs in achieving a real-time integrated virtual environment simulation with an extremely high level of detail. These challenges mainly concern physical characteristics and physical phenomena, the behavior of other road users, and noise characteristics for sensors and vehicle motion.
Hence, one of the significant challenges is replicating a realistic sensor input to the vehicle under test. 
Synthetic signals are commonly utilized in commercial \cite{CarMaker,VTD,Prescan,Drive_Sim} and Open-Source simulators \cite{Dosovitskiy17,LGSVL}. 
Their fidelity is affected by both the quality of the sensor model and the quality of the virtual environment model.
Additionally, sensor models are computationally expensive, as synthetic signals potentially increase in size and details. For example, a synthetic camera must generate images at the exact resolution and rate of the modeled physical camera.
Therefore, this approach does not scale well due to hardware requirements.

However, synthetic signals are also a promising approach to solve other problems, such as data generation for machine learning tasks.
For example, Cortes et al. \cite{cortes2020analysis} have exploited synthetic signals to develop perception algorithms that perform better on real data.
However, Talwar et al. \cite{talwar2020evaluating} trained and tested perception algorithms on synthetic and real datasets. They have demonstrated that the trained models do not transfer well between real and synthetic datasets.

\section{Perception Evaluation and Modeling} \label{ssec:criticalities}
\noindent Since we aim to test S\&P systems, we present state-of-the-art perception evaluation metrics in this Section. Next, we investigate related works on sensor modeling.

In the Computer Vision (CV) field, perception software is often evaluated as an independent component.
We can divide perception errors into four categories: 
\begin{itemize}
    \item \textbf{Detections}: Detecting the surroundings can be affected by false positives and false negatives;
    \item \textbf{Misclassification}: Even if detected, any object can be associated with the wrong class.
    \item \textbf{Object parameters}: Any parameter can be potentially affected by measurement errors. Physical parameters with a continuous domain (e.g., size, position) inherently involve a degree of precision and accuracy. 
    \item \textbf{Dynamics}: Each error can evolve. Detected objects can be lost (tracking error), a class label could be changed. Furthermore, object parameters can physically change (e.g., moving objects), making the error dynamic.
\end{itemize}
\subsection{Evaluation Metrics}
By comparing the ground truth and the CV algorithm output, we can measure the magnitude of all error types and summarize them into a single metric.
CV algorithms developed for generic object detection tasks are often evaluated by Intersection over Union (IoU) and Mean Average Precision (mAP) metrics \cite{Everingham2010,Cordts2016Cityscapes}. 
Similarly, tracking algorithms are evaluated with Multiple Object Tracking Accuracy (MOTA) and Multiple Object Tracking Precision (MOTP) \cite{Bernardin2008}.
These metrics are developed without considering the deployment context of the algorithms so that they apply to any domain.
Thus, these metrics are not aware of the composition of the scene and hence are context independent. However, for AV applications, the composition of the urban scene affects whether specific errors (e.g., misdetections) are critical.

We observe that the most desirable property of S\&P is to provide \textit{sufficient} information to DP to afford a safe reaction to the surrounding. A perception error is critical only if it leads to unsafe behavior, and it is negligible otherwise.
For example, while undesirable, misclassifying an obstacle on the road is not critical if the AV stops or avoids it properly. On the other hand, the same error in a different context may lead to an unacceptable response or even a collision.
Addressing this limitation, studies such as \cite{cheng2018towards,lyssenko2021evaluation,volk_comprehensive_2020,cheng2021safety}  propose metrics that integrate context-specific criteria in the evaluation.

\begin{figure}[t]
     \centering
     \includegraphics[width=.9\columnwidth]{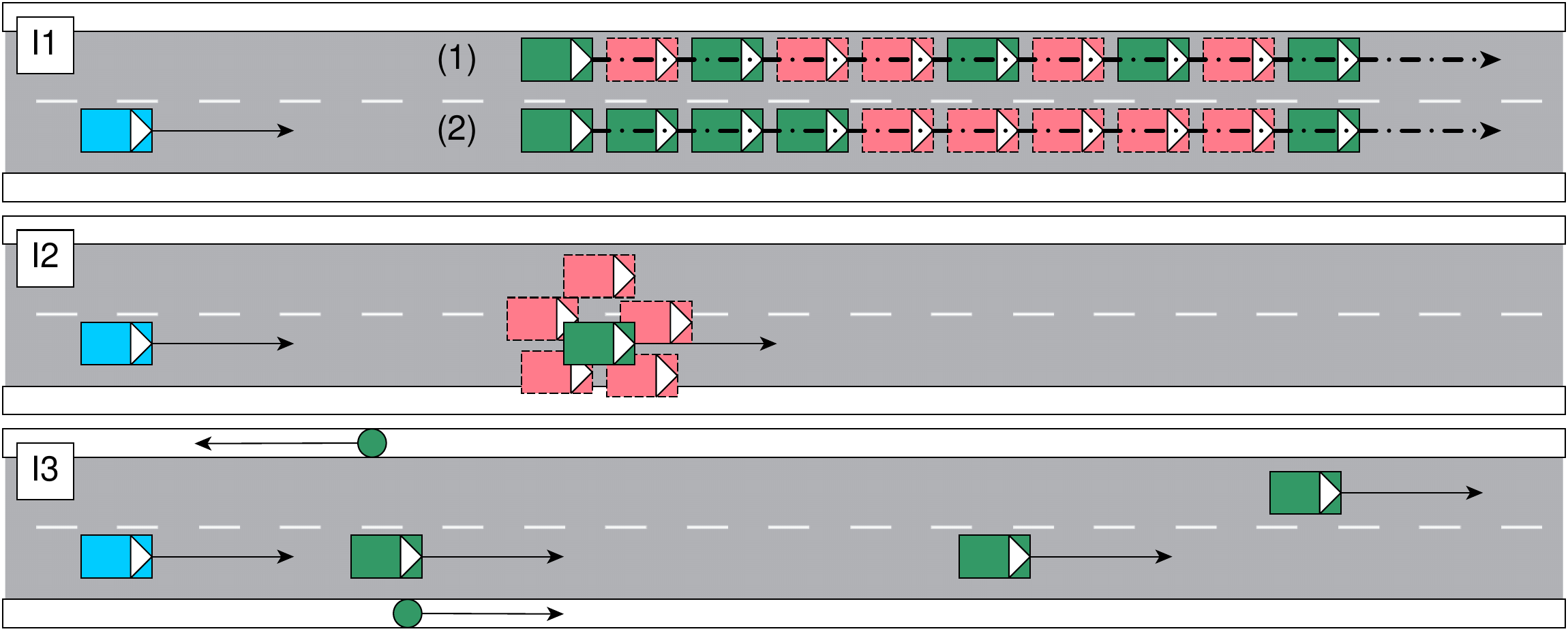}
     \caption[]{Illustration of critical issues I1, I2, I3. We indicate the ego vehicles in blue, other objects in green, and detections in red.
     (I1) Temporal Relevance: The detection timeline of the vehicle in lane 1 is more sporadic and unstable, but non-detection intervals are shorter than the vehicle in lane 2. 
     (I2): Overlap Sensitivity: A detection bounding box can be closer, farther, or on the side relative to the actual vehicle.
     (I3) Relevance of the objects: Many objects are present.
     In each of these examples, the context of the error affects its severity.}
     \label{fig:issues}
\end{figure}

We highlight four aspects regarding the connection between perception metrics and overall AV safety (see \autoref{fig:issues}), which are out of the scope of metrics derived from IoU and mAP:
\begin{itemize}
    \item \textbf{I1: Temporal Relevance}: AVs operate in a dynamic environment, where the state changes continuously and decisions are time-sensitive. The same error, e.g., misdetection, can have a different impact on safety based on the time interval it manifests on and how it evolves. Long intervals of misdetection may generate slightly higher safety concerns than sporadic episodes. 
    \item \textbf{I2: Overlap Sensitivity}: When detected, the coordinates of an obstacle are analyzed to determine the course of action (e.g., avoid, follow at a distance).
    A smaller magnitude of spatial errors is desirable, but limiting the evaluation to the bounding box overlap (IoU) alone may not be informative in terms of safety.
    \item \textbf{I3: Relevance of the objects}: 
    While driving on the road, AVs are surrounded by many objects, road users, traffic signals, and traffic lights. At any specific time, only a subset of these objects is relevant to DP.
    Furthermore, the relevance varies over time, as relative distances, directions, and the potential of a collision change continuously.
    \item \textbf{I4: Impact on Driving Policy}:  Perception Modules deployed in AVs directly provide their output to the Driving Policy. Hence, the strengths and weaknesses of the two components are closely tied together, and evaluating the first without accounting for the second is shortsighted. For example, a DP may react poorly to misclassifications and robustly against false positives, while another may behave oppositely.
\end{itemize}
While novel, safety-aware perception metrics could address the first three considerations, no metric can be easily \textit{generalized} to any arbitrary Driving Policy,
as different driving styles may have different demands in perception quality. 
Thus, we approach this problem by employing virtual testing where we directly integrate the DP in the pipeline.

\subsection{Modeling Errors} \label{sec:relatedWorks}

\begin{table}[t]
    \centering
        \caption{Related work on Sensor Modeling.}
    \label{tab:sotaModeling}
    \begin{tabular}{c|c|c|c}
    Author &Method& Sensors &Dataset \\
    \hline
         Arnelid et al.\cite{arnelid2019recurrent}&RC-GAN & --  &Private\\
          Berkhahn et al. \cite{berkhahn2021traffic} & RODEs & -- &  \\
        Hirsenkorn et al. \cite{hirsenkorn2016virtual} & Non-parametric & Radar& Private \\
        Krajewski et al. \cite{krajewski2020using} & NN & LiDAR & Private\\
        Mitra et al.
         \cite{Mitra2018} & RNN & Camera & KITTI \cite{Geiger2012}\\
        Innes et al. \cite{innes2022testing} & NN & -- & KITTI \cite{Geiger2012} \\
         Sadeghi et al. \cite{sadeghi2021step} & NN& -- & Synthetic \\
        Zec et al. \cite{zec2018statistical} & AIOHMM & -- & Private\\
        Zhao et al. \cite{zhao2020method} & LiDAR equation & LiDAR & Private\\
    \end{tabular}

\end{table}
Sensor error (or noise) modeling is a valuable technique in signal processing. In fact, by applying integration algorithms, it is possible to drastically increase the sensors' viability. For example, Kalman filters assume Gaussian noise and linear dynamical models.
In our context, we are interested in modeling the error \textit{after} the perception layer, i.e., at the object level.
Hoss et al. \cite{hoss2021review} provide an extensive review of the state-of-the-art of this specific task. The authors highlight the lack of shared procedures and datasets.

In \autoref{tab:sotaModeling}, we summarize recent works on sensor error modeling. 
Mitra et al. model 2D bounding boxes errors in a camera-based system by learning Recurrent Neural Networks \cite{Mitra2018}. Conversely, Hirsenkorn et al. model distance measuring error of a RADAR-based system employing a non-parametric approach  \cite{hirsenkorn2016virtual}. We note that the different nature of the models makes quantitative comparisons not only unfeasible but also meaningless. 
Even if targeting the same sensor, Krajewski et al. \cite{krajewski2020using}, and Zhao et al. \cite{zhao2020method} follow two very different approaches to LiDAR modeling. 
Krajewski et al. \cite{krajewski2020using} model the sensor error as the difference between the object states \cite{krajewski2020using}. Notably, they employ drones to collect the Ground Truth, which is then compared to the object states perceived by a LiDAR-based system.
Zhao et al. \cite{zhao2020method}, instead, model the LiDAR physics to compute the position of each point in the point cloud. Nevertheless, the authors also approach the object level by providing a \textit{target mode}. Instead of the whole point cloud, 
they generate a single point for each object in the scene. This solution provides a slightly faster implementation, as the size of the synthetic data each step generates is drastically reduced. However, this approach cannot easily include other sensors, as the model is based on LiDAR physics. Moreover, the model does not imply a perception layer with fusion and tracking in the pipeline.
To address this aspect, Zec et al.~\cite{zec2018statistical} and Arnelid et al.~\cite{arnelid2019recurrent}, propose sensor-agnostic models. 
Both of their model, Autoregressive Input-Output Hidden Markov Model (AIO-HMM)~\cite{zec2018statistical} and Recurrent Conditional Generative Adversarial Networks (RC-GAN)~\cite{arnelid2019recurrent}, model object parameters such as location and size with time dependencies. These solutions are more suitable for AV virtual testing since AVs rely on multiple sensors of diverse natures. The authors can also compare their results, as they analyze the same private dataset and have access to both models.
More recently, Sadeghi et al. \cite{sadeghi2021step} adopt PEMs proposed in our previous work \cite{ijcai2020-483}. Their implementation relies on a Neural Network trained on synthetic data and a LiDAR-based object detector. Moreover, they demonstrate the efficiency and effectiveness of using PEMs as surrogates for S\&P systems using a collection of scenarios from the CARLA Autonomous Driving Challenge \cite{CarlaChallenge}.
Innes et al. \cite{innes2022testing} also employ PEMs in a simulation study to estimate rare failure probabilities in automated emergency braking scenarios. The authors argue that a photorealistic simulation is not necessary to study the AV behavior as long as the input to the DP is affected by the same error distribution as the real one. 
Berkhahn et al. \cite{berkhahn2021traffic} conducted a simulation study injecting errors, modeled with random ordinary
differential equations (RODEs), in the simulation pipeline to study their impact on intersection conflicts. Interestingly, the perception errors modeled are intended for human drivers and not for an S\&P of an AV.

In summary, various approaches for modeling perception errors have been proposed in the literature, such as probabilistic models, neural networks, non-parametric models, or physics-based models.
However, we note that most of them rely on private datasets, preventing a direct model comparison. 
Furthermore, most of these studies are still closely tied to specific sensors (e.g., camera, LiDAR) and model them individually.
However, Hanke et al. present a modular architecture where individual sensor models are combined and integrated into a simulation pipeline \cite{hanke2016classification}.

In contrast to all those existing approaches from the literature, we model the S\&P as a whole, including the sensors and the AI processing of the sensor data. Our approach is not limited to specific sensors or AI algorithms, since we model errors in a generic way that applies to any particular perception system. Thus, our approach shifts the perspective from sensor models to perception error models (PEMs).

Moreover, existing studies have not yet introduced a comprehensive and flexible pipeline for simulation experiments.

\section{Perception Error Models}\label{sec:3erMod}
\noindent In this Section, we propose a generalized definition of Perception Error Models. We propose a sensor-agnostic definition, which abstracts from the modeled sensors and, thus, has a standard interface. 
This assumption is helpful for flexible integration in a simulation pipeline, which would greatly benefit the field if adopted. Additionally, we summarize some general considerations for error modeling.

\subsection{Generalized Notation and Definition of PEM}
\noindent We denote the collection of $n$ surrounding Ground Truth (GT) objects (obstacles or road users) $\W = \{o_1, \ldots ,o_n\}$. The S\&P observes the ``world'' $\W$  and generates the perceived world $\OM = \{\tilde{o}_1, \ldots ,\tilde{o}_m\}$, i.e., the collection of the $m$ perceived objects.
Next, the Driving Policy DP generates a response $R$ by analyzing $\OM$.
Real objects $o$ and perceived objects $\tilde{o}$ are represented with an array of parameters.
Depending on the S\&P system, this array may include the class (e.g., vehicle, pedestrian) and the 3D bounding box of the objects, described by its location (x,y,z), orientation (yaw, pitch, roll), and dimensions (length, width, height).
Moreover, an object may also include parameters such as velocity, brake and turning indicator status, or even an activity label (e.g., parking, running, crossing). 
Depending on the system, we can extend this notation to include other road elements or a more refined classification that discriminates between cars, bikes, trucks, or other classes.
Generally, some S\&P systems may assume a flat surface, especially if the system cannot measure the depth or simplify vehicle dynamics. Such assumptions reduce the number of relevant parameters, e.g., assuming pitch and roll equal to $0$.

In \autoref{fig:AV_pipeline}, we note $\text{RawData} = \text{S}(\W)$ and $\OM = \text{P}(\text{RawData})$. We can observe that: 
\begin{equation}\label{eq:W+error}
        \OM = \text{P}(\text{S}(\W)) = \text{S\&P}(\W) = \W + \E,
\end{equation}
where $\E$ is the error between $\OM$ and $\W$.
The response $R$ determines the AV \textit{behavior} and, therefore, the overall safety and performance of the autonomous system:
\begin{equation}\label{eq:Dp+error}
        R = \text{DP}(\OM) = \text{DP}(\W + \E).
\end{equation}

In this article, we define Perception Error Model (PEM) as an approximation of the combined function of  sensing subsystem S and a perception subsystem P:
    \begin{equation}\label{eq:pemFunction}
    \text{PEM}(\W) \approx \text{S\&P}(\W) = \OM = \W + \E. 
    \end{equation}
This definition outlines the interface and role of the model. A PEM receives the ground truth world $\W$  and returns the perceived world $\OM$.

\subsection{Error Modeling Considerations}
\label{ssec:err_model_considerations}
We outline four key factors that alter the manifestation of perception errors $\E$ in AV, which should be considered while designing a PEM.
\paragraph{Positional Aspects} The relative position of $w$ w.r.t. the ego-vehicle slightly affects the quality of the S\&P response \cite{rosique2019systematic}. Sensors have limited Field of View (FoV), e.g., limited range, and blind spots. The relative distance also affects the signal strength and resolution.
Furthermore, S\&P systems usually perform better in areas where the FoV of multiple sensors overlap, as the signals from the sensors might be fused to generate the perceived world $\OM$.
\paragraph{Parameter Inter-Dependencies} 
Object parameters $\mathbf{X, C}$ are clear players in $\E$ manifestation \cite{Hoiem2012}.
For example, larger objects are more likely to be detected, slow-moving vehicles are easier to track, or classification errors affect size estimation.
Additionally, even parameters not relevant to DP may be crucial, such as the color or material of the object \cite{rosique2019systematic}.
For example, LiDAR and radar have degraded performances on dark/non-reflective surfaces or metallic objects. 
\paragraph{Occlusion} Objects on the road may influence the detection of other objects. For instance, large objects (e.g., trucks) may occlude smaller objects (e.g., cars or cyclists).
Occlusion has much impact on the reliability of perception systems.
Cumulative statistics computed over both occluded and non-occluded objects may yield a biased assessment of the actual performances of an S\&P module. It is recommended to report performance statistics for different levels of occlusion. 
\paragraph{Temporal Aspects}
Objects often move in the scene. For example, a previously occluded object may become not occluded, and vice versa.
Additionally, algorithm uncertainties and filtering techniques are affected by their previous state.
Hence, the error $\E$ affecting an S\&P also evolves.

\subsection{Validation}
The objective of PEMs is to inject perception errors into the simulation pipeline. Thus, their validity is achieved if and only if the AV behavior generated embedding a PEM is comparable to the behavior the same AV would generate employing an actual S\&P.
Sadeghi et al. \cite{sadeghi2021step} use synthetic data to tune their model. Thus, they can directly compare the resulting behaviors, as the sensing module (synthetic) and the vehicle dynamics (simulator) are exactly the same in both approaches.
However, if the PEM is tuned on real data, a proper validation would require physical testing to validate the PEM and the vehicle dynamics embedded in the simulator.
Thus, a more straightforward but arguably less effective validation is ensuring that the errors have the same magnitude and distribution as the S\&P that the PEM is replacing. 
\begin{figure}[t]
\centering
 \includegraphics[width=\columnwidth]{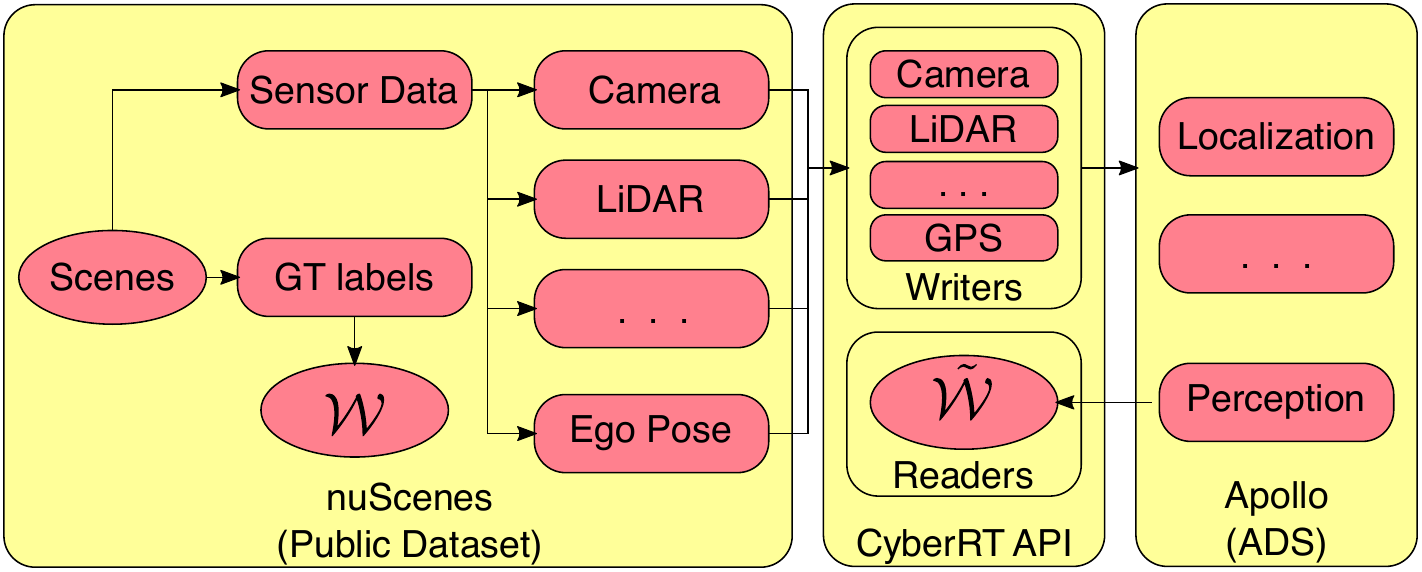}
\caption{Software pipeline to generate $\OM$ by means of Apollo \cite{Fan2018} and nuScenes \cite{Caesar2020}.
}\label{fig:makeOM}
\end{figure}

\section{Data Driven PEM}

\noindent As mentioned in \autoref{sec:relatedWorks}, most studies rely on private datasets. Our approach, instead, relies on a public dataset and open-source algorithms.
We define a perception dataset as a collection $S$ of scenes $s$. Each scene is a stream of pairs $( \W^s_t,\OM^s_t )$, where $t=1,2,\ldots T$ is discrete time.

To generate this dataset, we have considered several CV datasets, including KITTI~\cite{Geiger2012}, ApolloScape~\cite{huang2018apolloscape}, nuScenes~\cite{Caesar2020}, Lyft Level 5 dataset~\cite{lyft2019}, and Waymo open dataset~\cite{waymo_open_dataset}. 
Such datasets are mainly used to design perception systems (P), which are assessed by comparing the generated $\OM$ with the ground truth $\W$. By contrast, we consider an existing P, i.e., the open-source P system from Baidu Apollo \cite{Fan2018}, and design the PEM by modeling the error $\E$ as the difference between $\W$ and $\OM$.
We opted for nuScenes as the chosen example to illustrate our methodology due to the inclusion of RADAR data, rich annotations (used as ground truth $\W$), and data collected in Singapore and Boston.
Only a few implications derive from our selection of a public dataset and a perception algorithm. To avoid any bias, we highlight such implications.
The nuScenes choice guided us in the approach regarding the model of occlusion. This approach can be easily replicated on any dataset by computing the occlusion levels similarly. Nevertheless, a different model of occlusion can be explored. 
The choice of Apollo, instead, has a substantial implication on the implementation.
While modeling S\&P as a whole,  the PEM's output must follow the original perception module P output format.
With the described setup, the developed PEM models a hypothetical vehicle that employs the Apollo perception module P and the nuScenes sensors setup S, including a frontal camera, LiDAR, and RADAR.

\subsection{Perception Dataset}\label{ssec:PercDat}

\autoref{fig:makeOM} illustrates the procedure to generate the perception dataset.
The 1000 scenes composing nuScenes, ~20 seconds streams on recorded sensor data, are annotated with ground truth $\W$ at 2Hz. 
We convert $\W$ in the same format and frame of reference of $\OM$. 
Then we adapt the raw sensor data and prepare the message streams towards the Apollo perception module.
To exchange data between modules,
Apollo employs Cyber-RT \cite{cyberRT}, a ROS-like middleware.
Hence, we can utilize Cyber-RT APIs to define writer and reader nodes.
In particular, we prepared as many writers as per sensor type, each to synchronously replay raw data to replicate a nuScenes scene.
While replaying each scene, the Apollo perception module is activated and processes the raw data providing the output $\OM$.
A Cyber-RT reader logs to file the generated $\OM$.
Next, it is possible to compose the perception dataset using $\W$ from nuScenes annotations and the recorded $\OM$ for each scene, obtaining a stream of pairs $( \W^s_t,\OM^s_t )$ for each scene.

Lastly, any multi-object detection task requires solving a matching problem. In each frame, both $\W$ and $\OM$ are composed of multiple objects. Establishing a connection between items from one set to the other is necessary to measure any error.
Bernardin et al. provide a well-established solution
\cite{Bernardin2008} common in the CV field. Alternatively, Arnelid et al. \cite{arnelid2019recurrent} and Zec et al. \cite{zec2018statistical} relied on the solution proposed by Florback et al. \cite{florback2016offline}.
All these matching algorithms are based on similarity functions.
Thus, it is essential to highlight that any parameter and procedure defined in this task will alter our error models. In particular, we imposed a threshold of 10m as the maximum distance between $o_i$ and $\tilde{o_j}$ to allow a match.

\subsection{Learning PEMs}
In this section, we explain how to learn a PEM from data, using the dataset presented in \autoref{ssec:PercDat}.
Given the generalized PEM definition (\autoref{eq:pemFunction}), a PEM can be implemented in various ways, and the goal is to generate $\Tilde{\W}$ given $\W$.
For the scope of this article, we design our model to be easily relatable to basic evaluation metrics such as detection rates and IoU, as well as to provide a clear example of ways to address the considerations provided in \autoref{ssec:err_model_considerations}.

The PEM we propose operates in two steps when integrated into the simulation.
First, it processes each object $o_i \in \W$ and determines whether it is detected, generating the corresponding perceived object $\tilde o_i$. Second, it calculates the magnitude of error $\varepsilon$ on the parameters of $\tilde o_i$. The PEM then composes $\Tilde{\W}$ as the set of all $\tilde o_i$, and returns it to the simulator.
To model the detection of objects, we introduce a binary state variable $v$, where $v=1$ when the object is detected and $v = 0$ otherwise. We parameterize the objects by their location, expressed in polar coordinates (radial~$r$, angular~$\theta$), and their perceived values  as $(\tilde r, \tilde \theta)$.

\begin{figure}[t]
         \centering
         \includegraphics[trim=12cm 6cm 12cm 6cm, clip=true, totalheight=0.7\columnwidth, angle=90]{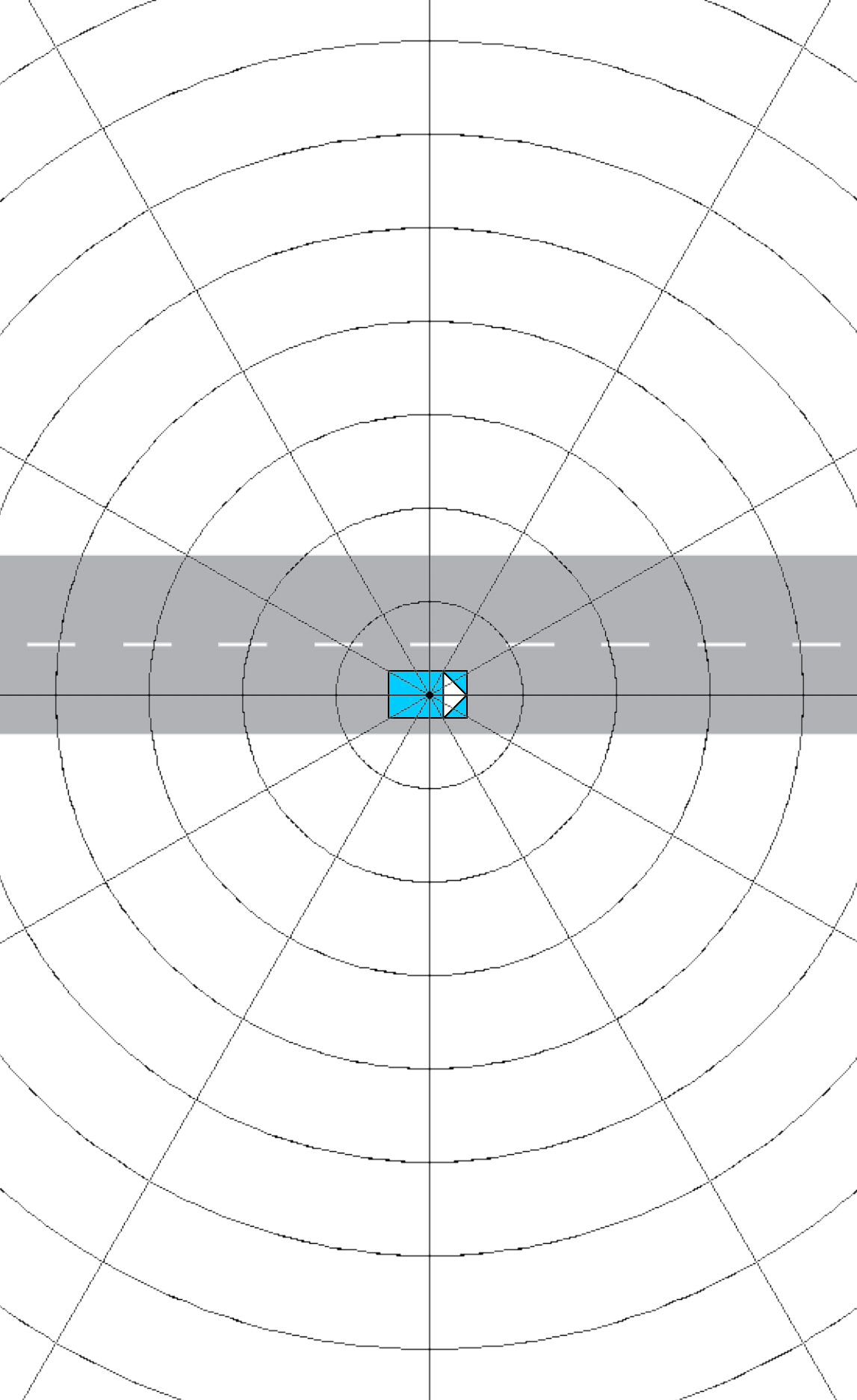}
         \caption{Example of a zone-based partitioning of PEM. A polar grid can meaningfully organize the surrounding area into zones, which different sensors may cover with a different degree of reliability.}
         \label{fig:polarGridExample}
\end{figure}

We model the evolution of the state $v$ by a Hidden Markov Model (HMM) with Gaussian emissions. In particular, a transition matrix $ \mathbf{A}$ models the state variable $v$, and the emission probabilities model the error $\epsilon$ on the object's parameters $(\tilde r, \tilde \theta)$. We highlight that the state variable $v$ is observed in the dataset, which greatly simplifies the learning task. Moreover, we do not introduce emissions for non-detected objects, since we do not need to generate errors on their parameters.
To account for the various factors that can affect the performance of the S\&P performance (see \autoref{ssec:err_model_considerations}), we propose dividing the model into several partitions based on external conditions $c \in C$.  Each partition is parameterized according to the relative data in the dataset. For example, a zone-based partitioning (as shown in \autoref{fig:polarGridExample}) generates partitions trained specifically on objects located in the corresponding zone.

We represent an input object $o_i$ in the scene with three variables $o_i = (r,\theta,c)$, i.e., the polar coordinates (radial~$r$, angular~$\theta$), and the corresponding values of $c \in C$. 
We describe perceived object $\tilde o_i$ with three state variables  $\tilde o_i = (\tilde r, \tilde \theta)$. With $\tilde \rho$  and $\tilde \phi$, we indicate the perceived polar coordinates, i.e., with error.  We use the notation $o_{it}$ to indicate the state of the object $o_i$ at time $t$.

We implement a PEM as a set of $|C|$ pairs:
\begin{equation}\label{eq:pem}
    \text{PEM} = \{(M^c, p^c), c \in C\},
\end{equation}
where:
\begin{itemize}
    \item $C$ is the set of external conditions $c$ that alter the perception quality. Following the considerations in positional aspects and occlusion, we define $C$ as the cross-product of a grid cell partition of the ego-vehicle surroundings and the set of occlusion levels. 
We can easily follow nuScenes occlusion modeling. The nuScenes dataset defines the four levels of visibility, i.e., the fraction of the annotation (i.e., the ground truth) visible from the ego-vehicle. It is binned into 4 bins: $0-40\%$, $40-60\%$, $60-80\%$, and $80-10\%$. Hence, our set $\C$ is generated as follows:
    \begin{equation}\label{eq:stept}
         \C = \{ \text{vis}_0,\text{vis}_1,\text{vis}_2,\text{vis}_3 \} \times \{\text{grid cells}\},
    \end{equation}  
where the polar grid is obtained with  $30\deg$ sectors and radius increments of $10m$.
    \item for each $c \in C$, $M^c$ is the Markov Chain that models $ v$, i.e., whether an object is detected. Given the considerations about temporal aspects, each $M^c$ is implemented by the transition matrix $\mathbf{A}^{c}$ to determine $ v$:
\begin{equation}\label{eq:transmat}
    \mathbf{A}^{c} = \begin{bmatrix}
a_{00}^c & a_{01}^c\\
a_{10}^c & a_{11}^c
\end{bmatrix},
\end{equation}
where the transition matrix parameters vary with $\C$, as $c$ refers to $o_{it}$ values.

The probability of an object $\tilde o_i$ at time $t$ is obtained via Maximum Likelihood Estimation (MLE):
\begin{equation}\label{eq:learnMC}
 P( v_{it} = l | v_{i,t-1} = k, o_{it})  =   \mathbf{A}^{c}_{kl} = \frac{w^{c}_{kl}}{\sum_{m\in D}{w^{c}_{km}}},
\end{equation}
where $w^{c}_{kl}$ are the transitions from state $v=k$ to $v=l $ that occur in $c \in \C $ in the dataset $( \W^s_t,\OM^s_t )$ (see from \autoref{ssec:PercDat}). 
    \item $p^c$ describes the parameter error $\varepsilon$ distribution, i.e., the difference between perceived and ground truth parameters:
    \begin{equation}
    \varepsilon = ( \varepsilon_{\tilde{r}}, \varepsilon_{\tilde{\theta}}),
    \end{equation}
    \begin{equation}
    \varepsilon_{\tilde r} = \frac{\tilde r}{ r}, \\
    \end{equation}
    \begin{equation}    
    \varepsilon_{\tilde \theta} =\tilde{\theta}-\theta.
    \end{equation}
    If object $o_{it}$ is detected, the object $\tilde{o_{it}}$ parameters $\left(\tilde r, \tilde \theta\right)$ are generated with error distributed as a multivariate Gaussian distribution $\mathcal{N}^{c}$ over the parameters $r,\theta$ of $o_i$:
\begin{equation}( \varepsilon_{\tilde{r}}, \varepsilon_{\tilde{\theta}}) \sim \mathcal{N}(\mu^{c} ,\Sigma^{c}),
\end{equation}
where $\mu^{c} = (\mu^c_r,\mu^c_\theta)$ and $\Sigma^{c}$ represent means and covariances on the error for objects $o_i$ with $c \in \C $.
\end{itemize}

This model is entirely sensor agnostic, as no parameter is tied to any specific sensor or sensor output (e.g., point cloud, 2D bounding boxes).
 
\begin{figure}[t]
\centering
\includegraphics[width=\columnwidth]{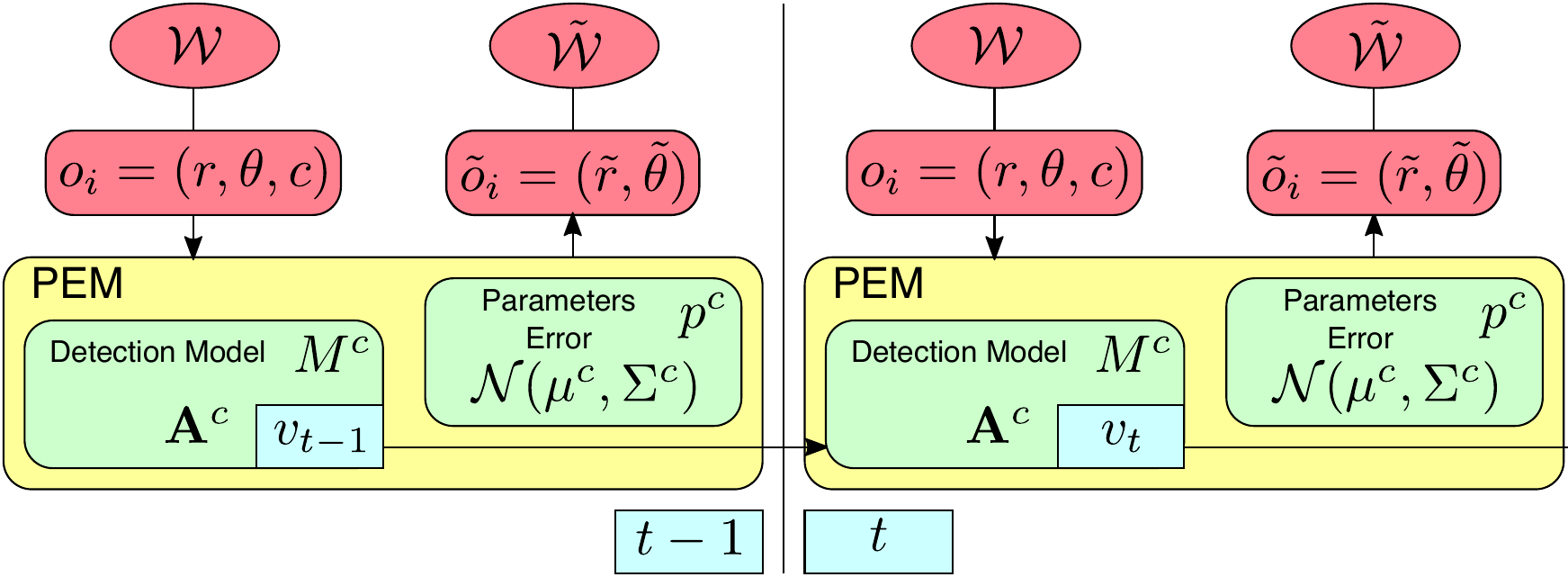}
\caption{Illustration of 2 steps (frames) of PEM we deploy in our experiments.
}\label{fig:pemDetails}
\end{figure}

Thus, we need to estimate 7 parameters for each $c \in C$. First, each model $M^c$ has 2 degrees of freedom. We can  estimate $a_{11}^c$ and $a_{01}^c$, and since each rows sum to $1$, we can derive $a_{10}$ $a_{00}$.
Second, each distribution $p^c$ has 5 parameters: the means $(\mu^c_r,\mu^c_\theta)$, and the parameters ($\sigma^c_r$ ,$\sigma^c_\theta$, $\sigma^c_{r,\theta}$) in $\Sigma^{c}$:
\begin{equation}
    \Sigma^{c} = \begin{bmatrix}
\sigma^c_r & 0\\
0 & \sigma^c_\theta
\end{bmatrix}\begin{bmatrix}
1 & \sigma^c_{r,\theta}\\
\sigma^c_{r,\theta} & 1
\end{bmatrix}\begin{bmatrix}
\sigma^c_r & 0\\
0 & \sigma^c_\theta
\end{bmatrix},
\end{equation}
where $\sigma^c_{r,\theta}$ is the correlation value.

Similarly to evaluation metrics, these parameters can be obtained analytically via MLE for each $c \in C$ independently. 
However, a partitioned model approach introduces two main drawbacks. First, independently estimating parameters in each partition reduces the amount of data considered.
Second, zone-based partitioning discounts prior knowledge of zone adjacency, which is important because objects are rarely located in only one zone.
This problem is typical in spatial data analysis, where the goal is to identify regional patterns (e.g., pollution levels). To this end, hierarchical models such as Conditional Autoregressive (CAR) models are often employed \cite{cressie2015statistics}.
CARs embed such zone adjacency knowledge in the parameter estimation.
In particular, they assume that the parameter $y_c$ we want to estimate (e.g., $\mathbf{A}^{c}_{kl}$, $\mu^{c}$) is conditioned to neighboring values of $y_{n,n\ne c}$.
The hierarchical nature of the model allows us to define the conditioning in a probabilistic manner:
\begin{equation}
    y_c | y_n,n\ne c \sim \mathcal{N}\left(\alpha \sum_{n=1}^{|C|} b_{cn}y_n, \tau_n^{-1} \right),
\end{equation}
where $\tau_n^{-1}$ is the precision at each partition, with prior $\tau_n \sim \Gamma(1,1)$.
We describe the adjacency structure in the matrix $b$. In particular, we consider a partition adjacent to the other partitions that share an edge in the zone-based partitioning and have the same occlusion level.

A priori, we assume that each partition's most probable value is the average of the neighbors' values. However, this approach also admits slightly different values, but only if the data is sufficient to support such a hypothesis.
Intuitively, the CAR model acts as a smoothing technique that considers the amount of data to determine the smoothing factor. For example, partitions with no samples will converge to the average of the neighbors' values. In contrast, partitions with a high number of samples will converge the value supported by the samples.

The adoption of CAR models makes the parameters no longer obtainable directly via MLE, since the partitions are no longer independent. Therefore, we employ the probabilistic programming framework PyMC3 \cite{salvatier2016probabilisticPYMC3}, and define a CAR model for each of the seven parameters of the PEM. Next, we apply the PyMC3 inference framework to fit the parameters employing the dataset $( \W^s_t,\OM^s_t )$. In particular, we apply the Maximum a Posteriori method to obtain point estimates of the PEM parameters.

\begin{table}[t]
\centering
\caption{List of PEMs obtained using three different sensor setups.}\label{table:setups}
 \begin{tabular}{c | l } 
 ID & Sensor configuration \\
  \hline & \\[\dimexpr-\normalbaselineskip+2pt]
 CAM & Only frontal camera data.\\ 
 LID & Only Lidar data, i.e., only the point cloud.\\
 FUL & Full setup, combining camera, LiDAR, and RADAR data.\\
\end{tabular}
\end{table}

\subsection{Examples}\label{sec:pemlearnRes}

During the generation of a perception dataset illustrated in \autoref{ssec:PercDat}, we can easily decide which data, e.g., images or point clouds, we feed to the Apollo perception module (see \autoref{fig:makeOM}).
For the experimental scope of this paper, we selected three different sensor setups, summarized in \autoref{table:setups}.
We select these three configurations as they often occur in practice. Moreover, in this way, we can test several PEMs.
In \autoref{fig:polarVisibility}, we show the most direct metric, i.e., detection rate, in the context of PEM under each of the chosen sensor setups. We present a diagram for each element in $\C$. Thus, we depict a polar grid for each occlusion value. Within each of the delimited zones, we can show how a single parameter or a metric varies.
Detection rates are the most informative and intuitive to visualize as they depict the effective Field of View of the S\&P.
In Figures \ref{fig:sp_Cf}, \ref{fig:sp_Lf}, \ref{fig:sp_Ff}, we observe how the CAM detection rate has a slower decay in the frontal cone, followed by FUL and LID. The fusion algorithm present in FUL improves on LID beyond the 30m range but only partially exploits camera data at larger distances.

\newcommand\sizescatter{.175\textwidth}
\newcommand\sizebar{.175\textwidth}
\newcommand{\rulesep}{\unskip\ \vrule height\sizescatter width .01mm}
\newcommand\myhspace{\hspace{1mm}}
\begin{figure*}[t]
     \centering
     \subfloat[][CAM: vis0.]{\includegraphics[width=\sizescatter]{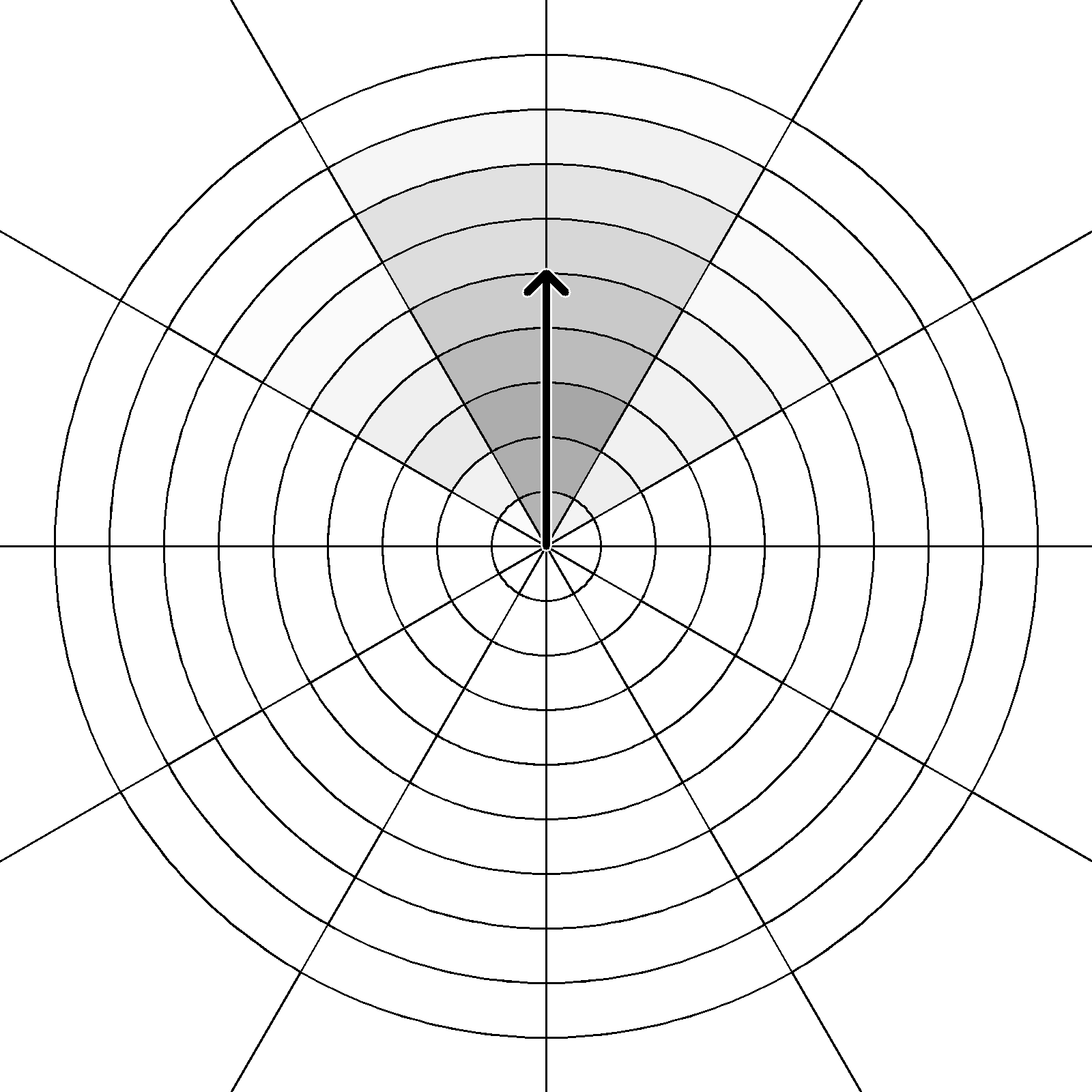}\label{fig:sp_Ca}}
      \myhspace
     \subfloat[][vis1.]{\includegraphics[width=\sizescatter]{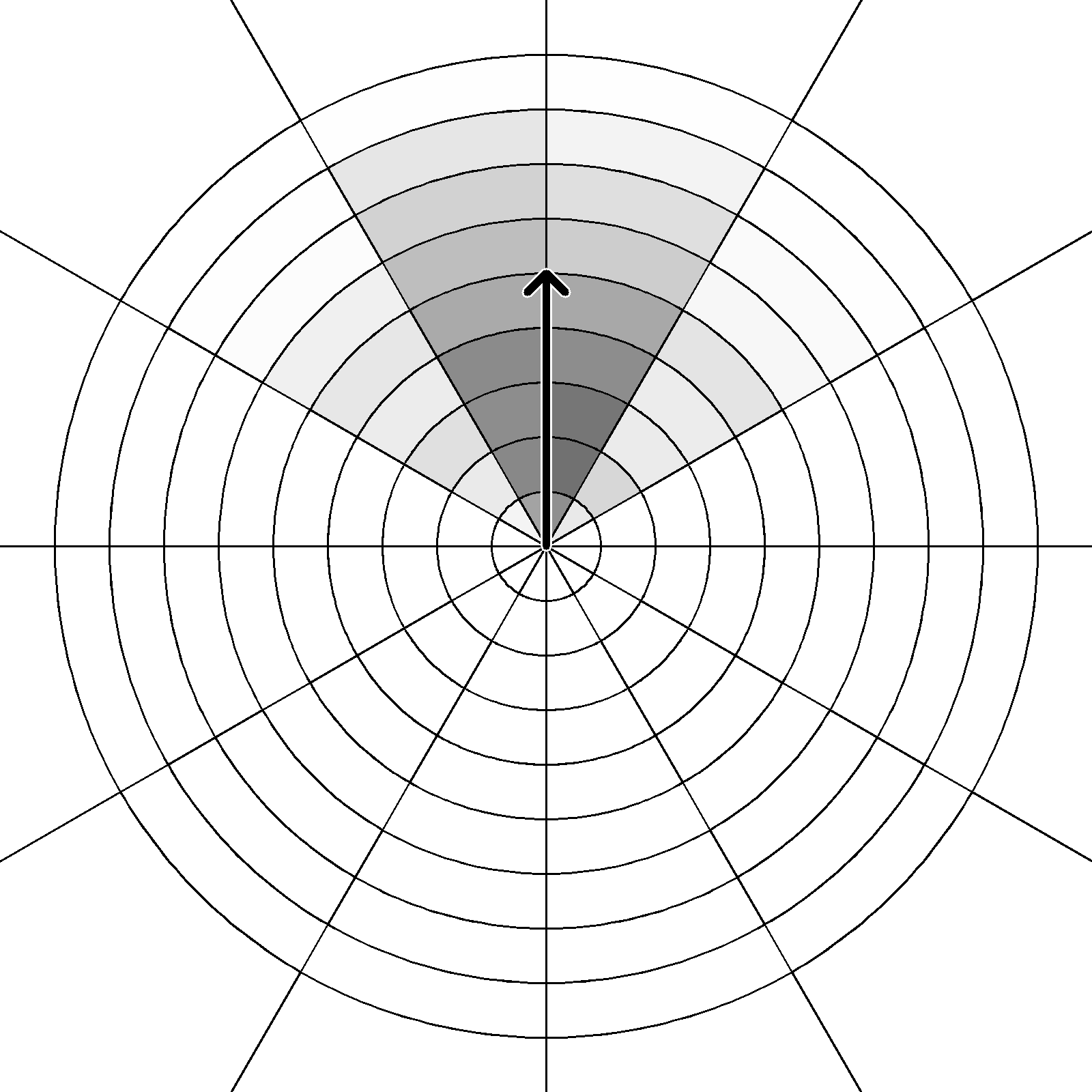}\label{fig:sp_Cb}}
      \myhspace
       \subfloat[][vis2.]{\includegraphics[width=\sizescatter]{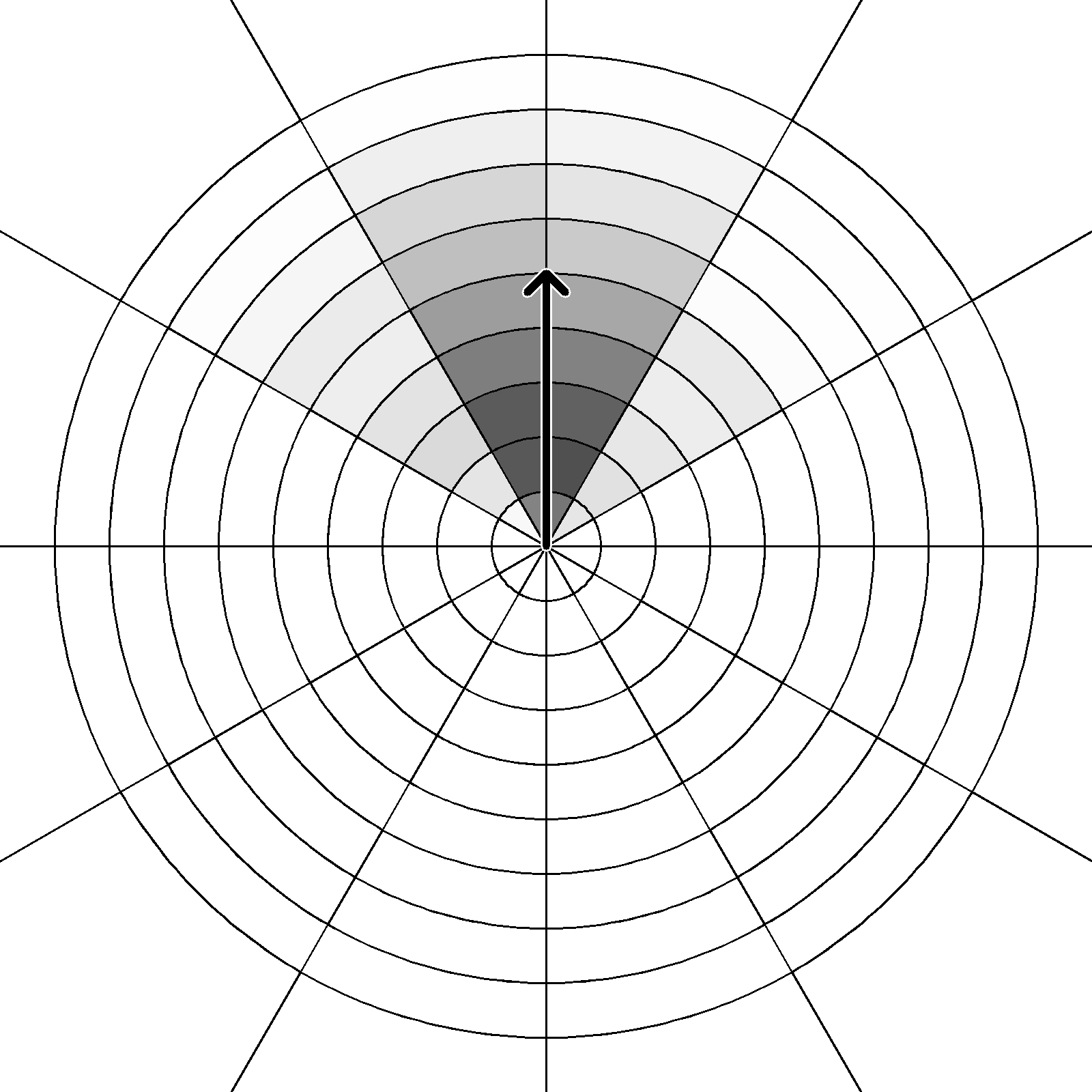}\label{fig:sp_Cc}}
      \myhspace
     \subfloat[][vis3.]{\includegraphics[width=\sizescatter]{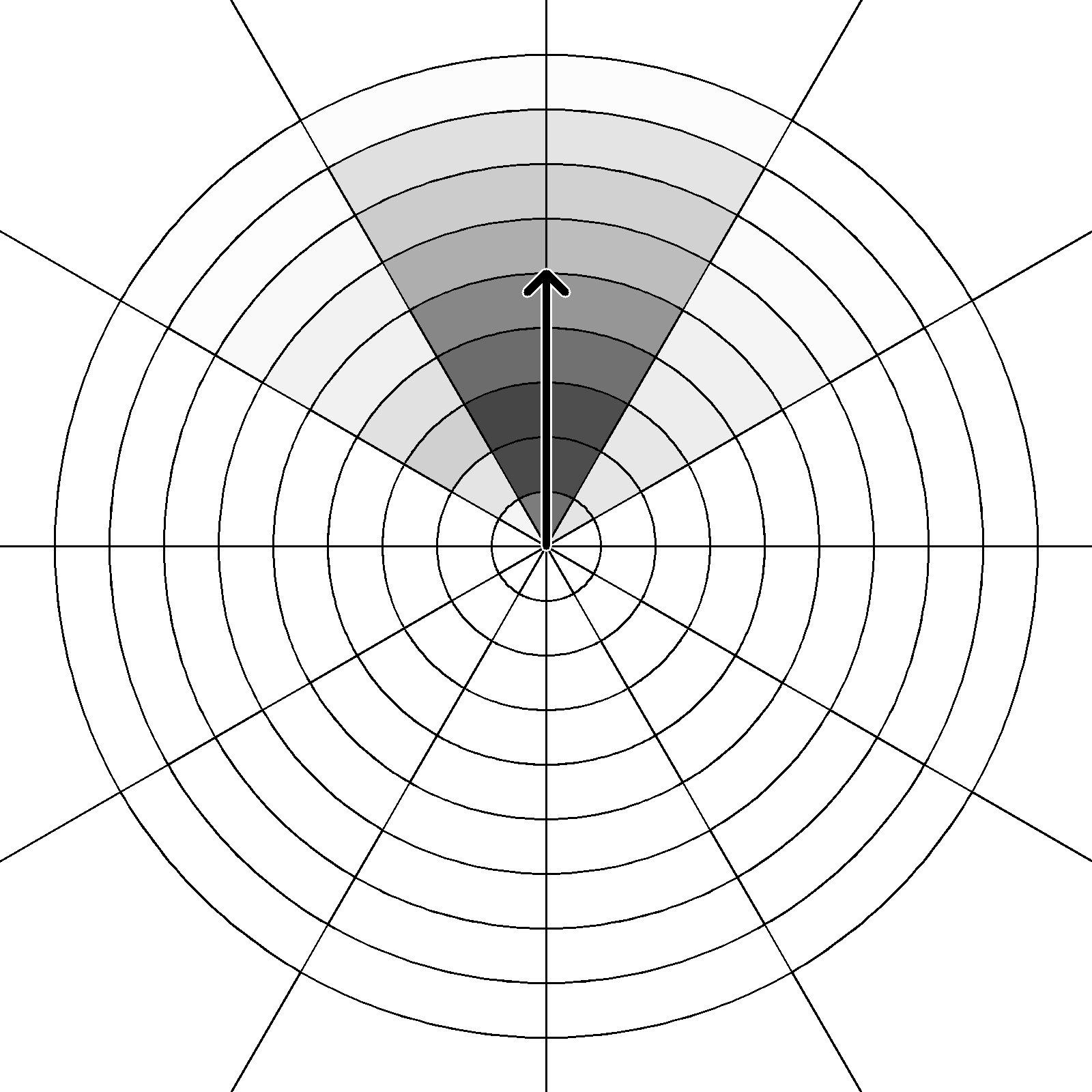}\label{fig:sp_Cd}}
     \myhspace
     \subfloat[][P]{\includegraphics[height=\sizebar]{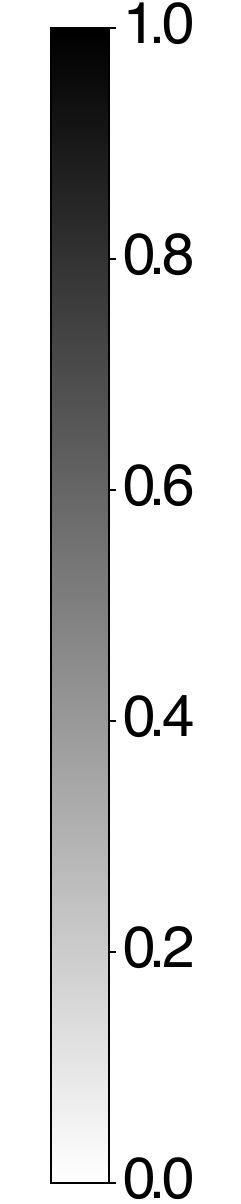}\label{fig:sp_Ce}}
     \myhspace
     \rulesep
     \myhspace
     \subfloat[][CAM Frontal cone P.\label{fig:sp_Cf}]{\includegraphics[width=\sizescatter]{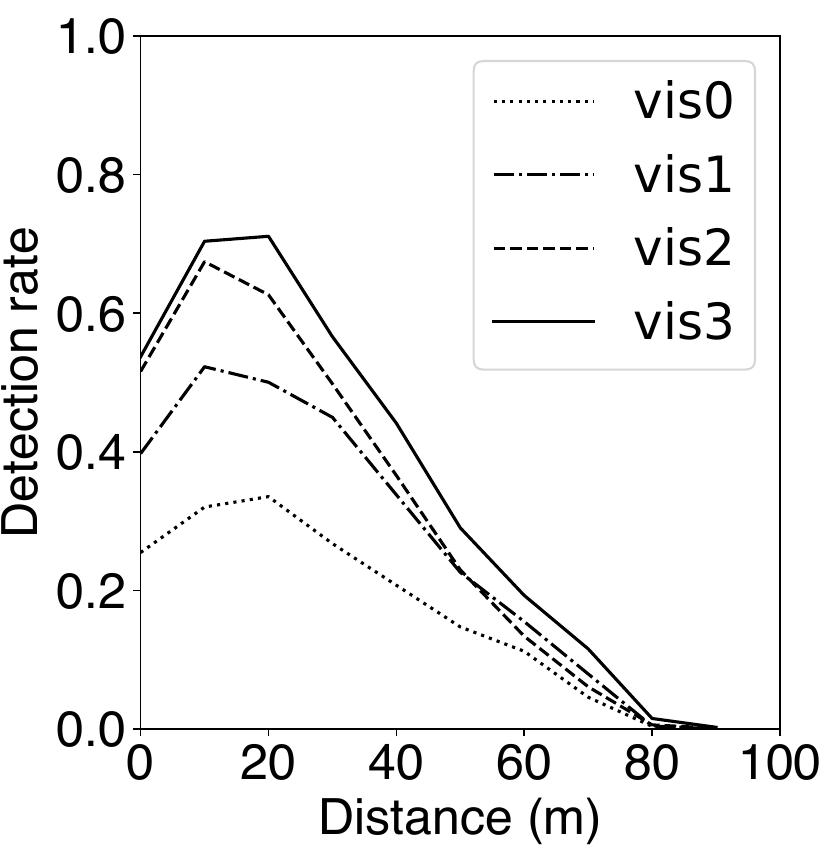}}     
     \\
     \vspace{-3mm}
      \subfloat[][LID: vis0.]{\includegraphics[width=\sizescatter]{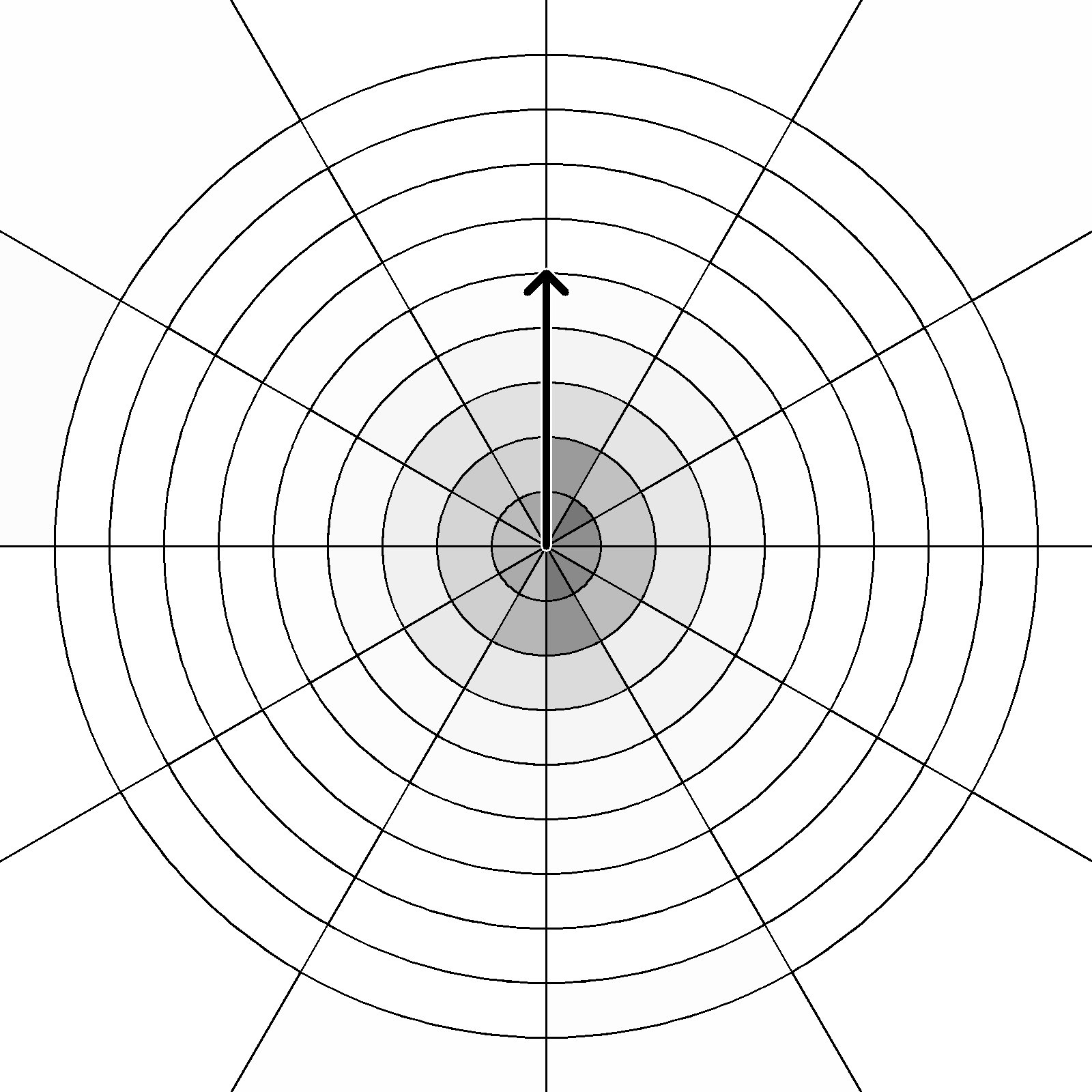}\label{fig:sp_La}}
      \myhspace
     \subfloat[][vis1.]{\includegraphics[width=\sizescatter]{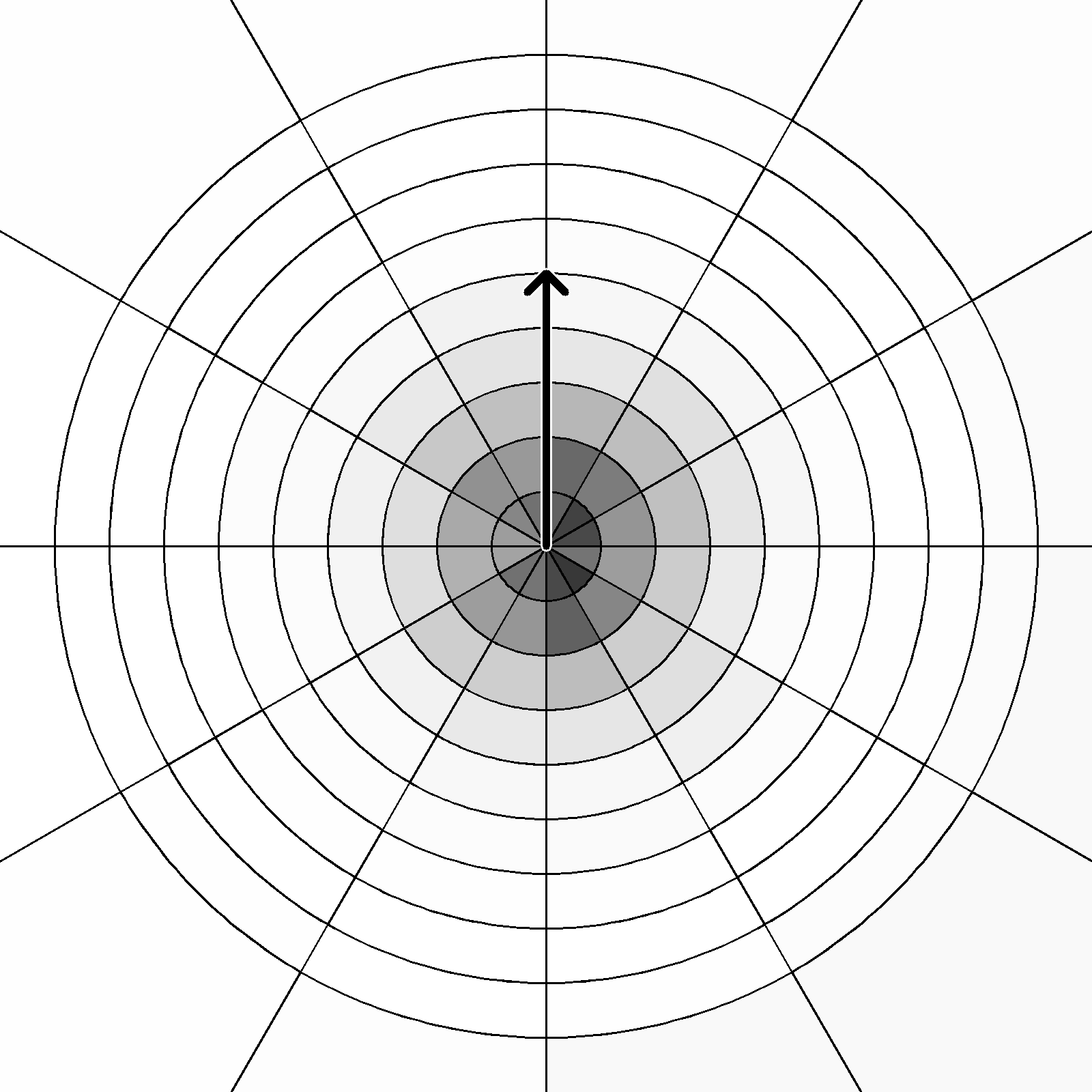}\label{fig:sp_Lb}}
      \myhspace
      \subfloat[][vis2.]{\includegraphics[width=\sizescatter]{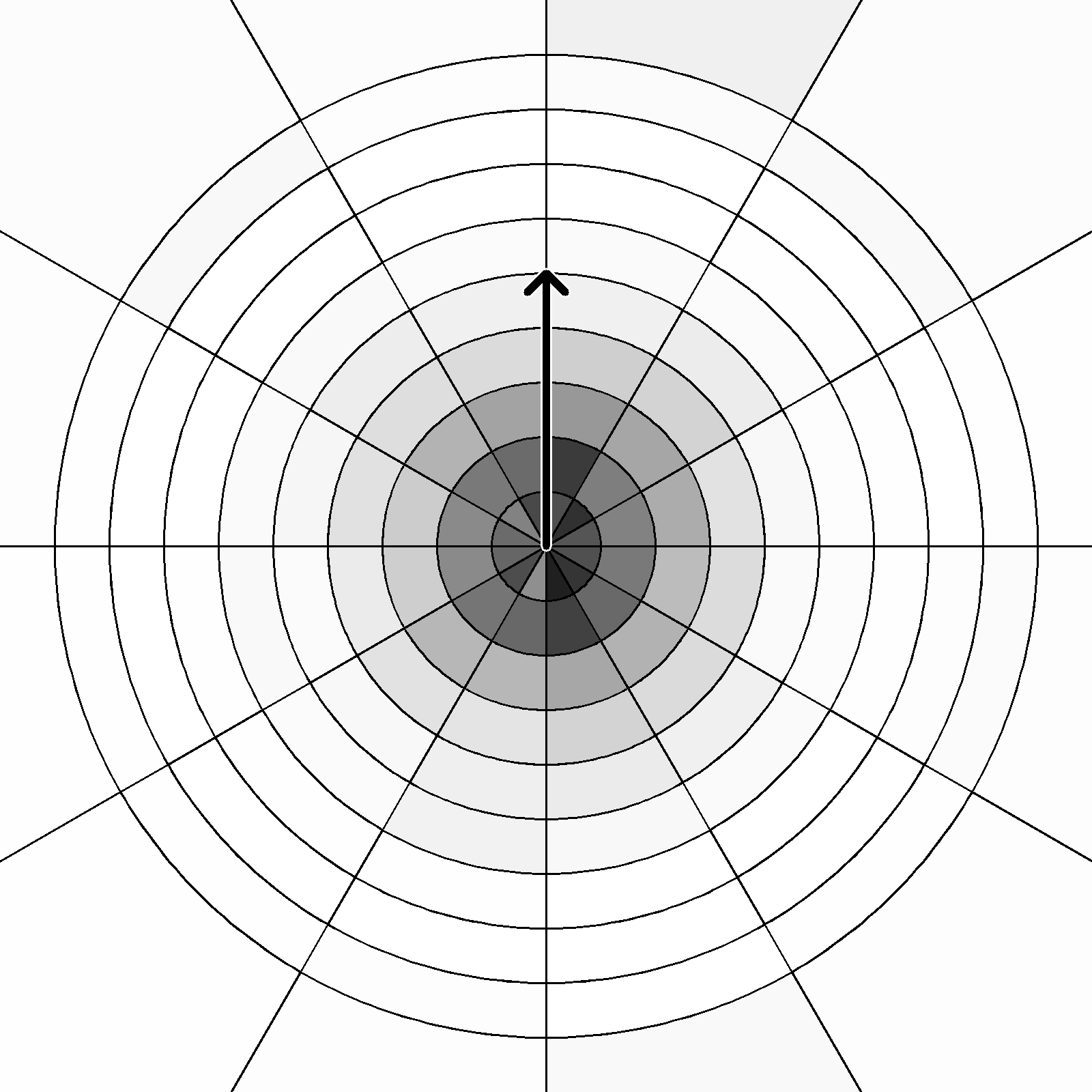}\label{fig:sp_Lc}}
      \myhspace
     \subfloat[][vis3.]{\includegraphics[width=\sizescatter]{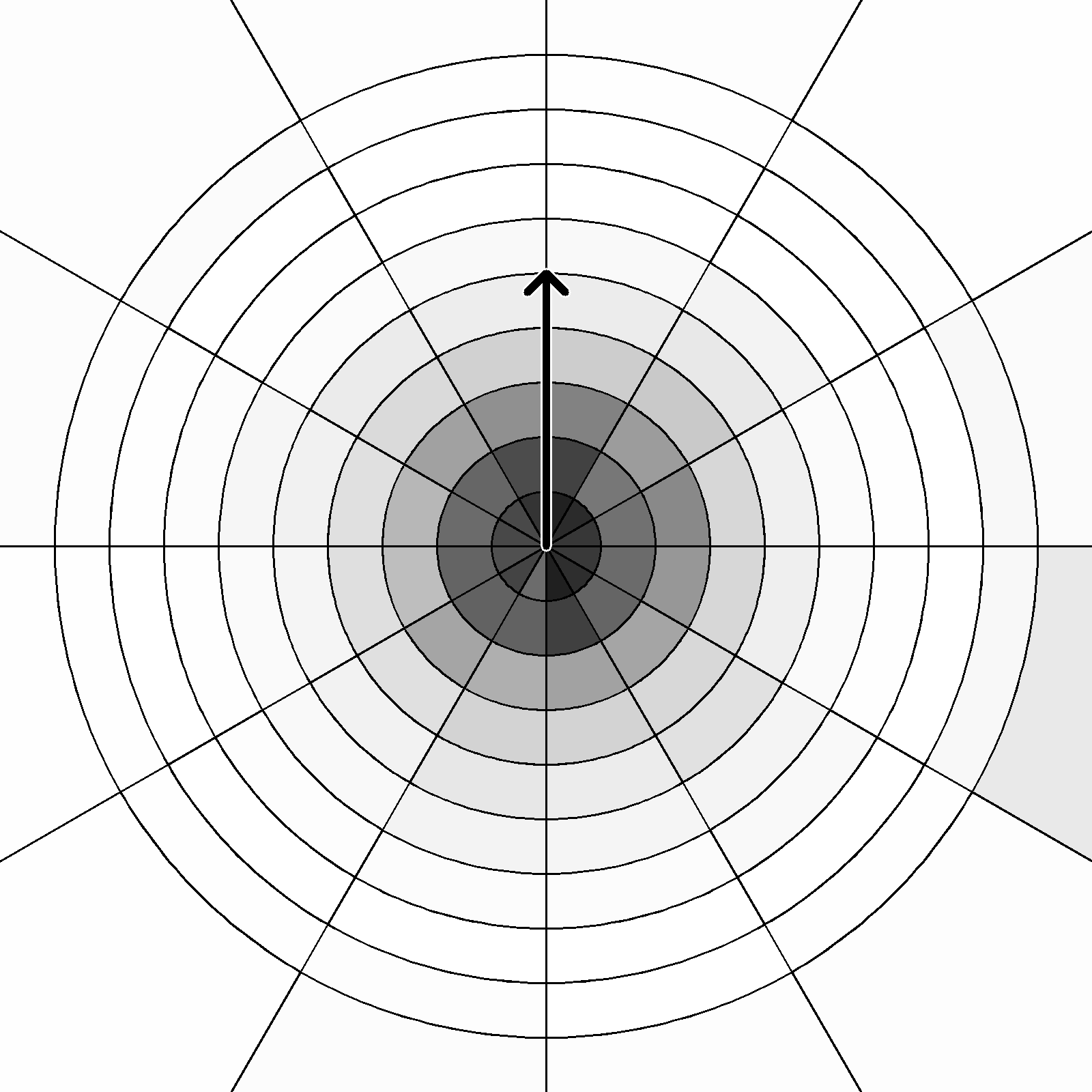}\label{fig:sp_d}}
     \myhspace
     \subfloat[][P]{\includegraphics[height=\sizebar]{figures/visibilityPlot/Figure_1.pdf}\label{fig:sp_Le}}
     \myhspace
     \rulesep
     \myhspace
     \subfloat[][LID Frontal cone P.]{\includegraphics[width=\sizescatter]{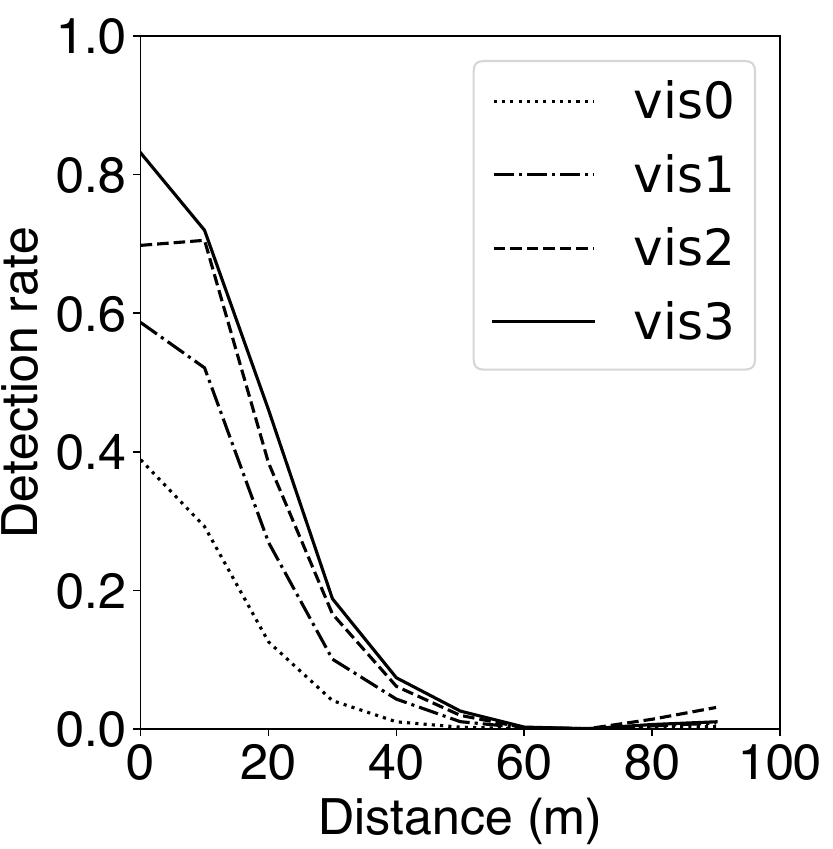}\label{fig:sp_Lf}}
     \\
     \vspace{-3mm}
     \subfloat[][FUL: vis0.]{\includegraphics[width=\sizescatter]{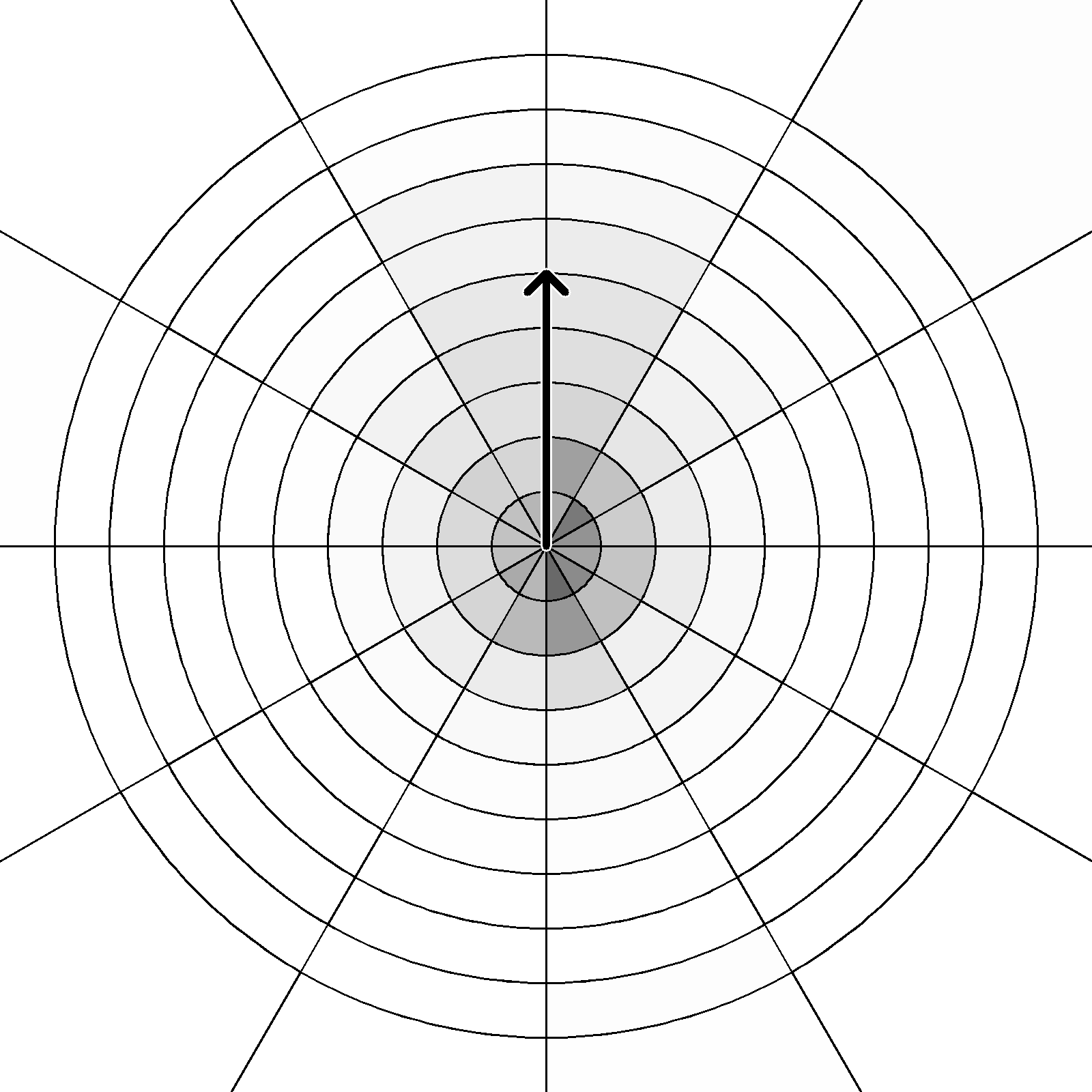}\label{fig:sp_Fa}}
      \myhspace
     \subfloat[][vis1.]{\includegraphics[width=\sizescatter]{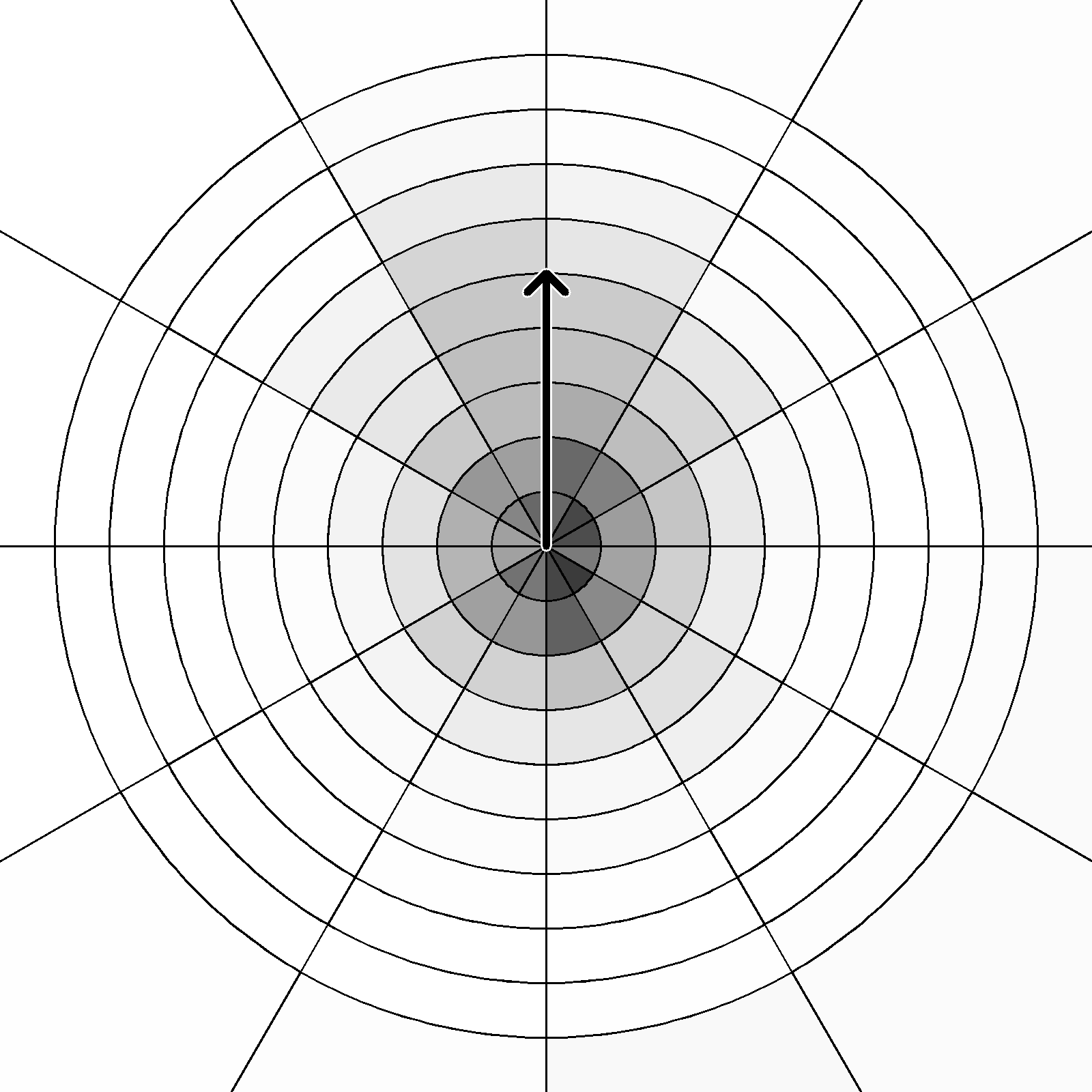}\label{fig:sp_Fb}}
      \myhspace
      \subfloat[][vis2.]{\includegraphics[width=\sizescatter]{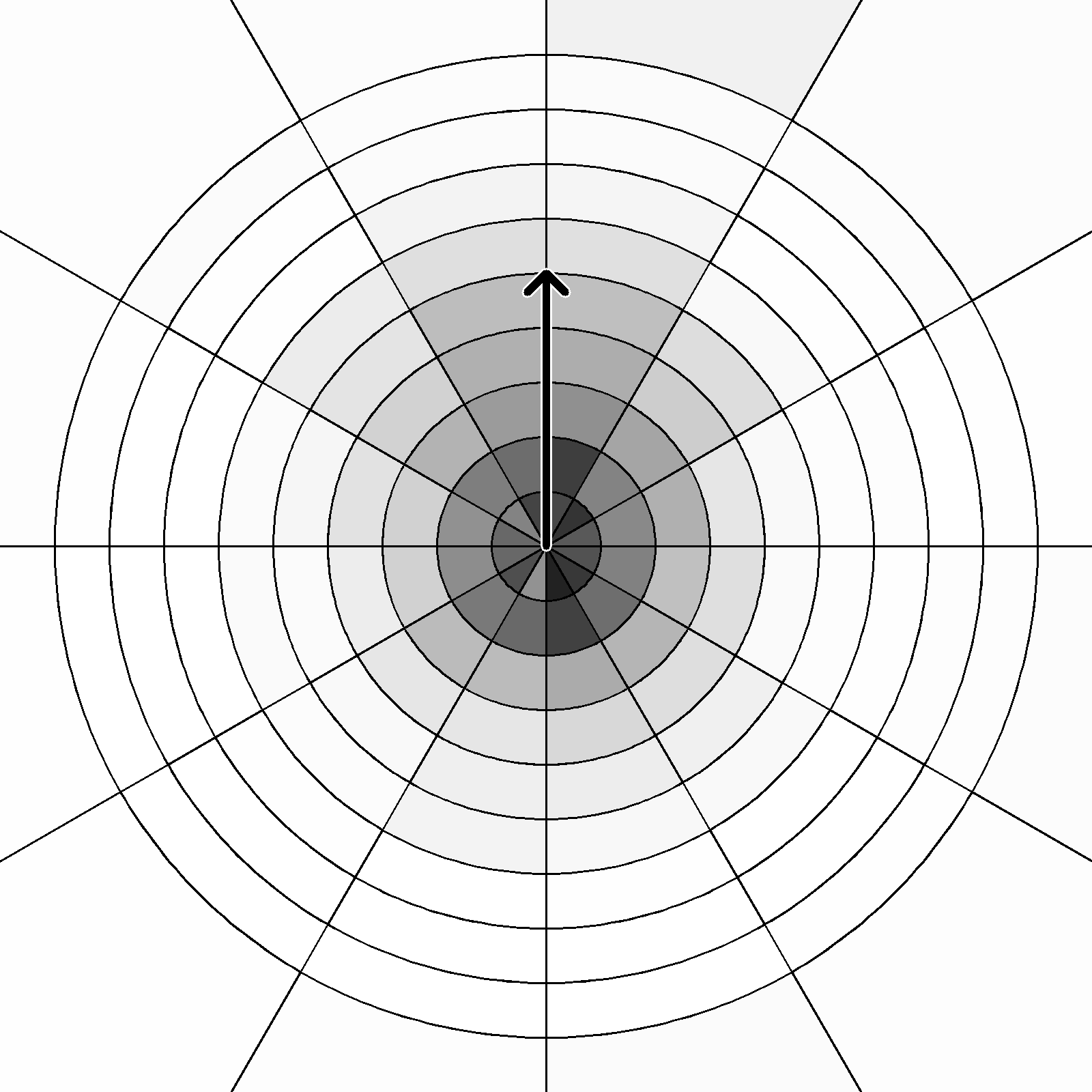}\label{fig:sp_Fc}}
      \myhspace
     \subfloat[][vis3.]{\includegraphics[width=\sizescatter]{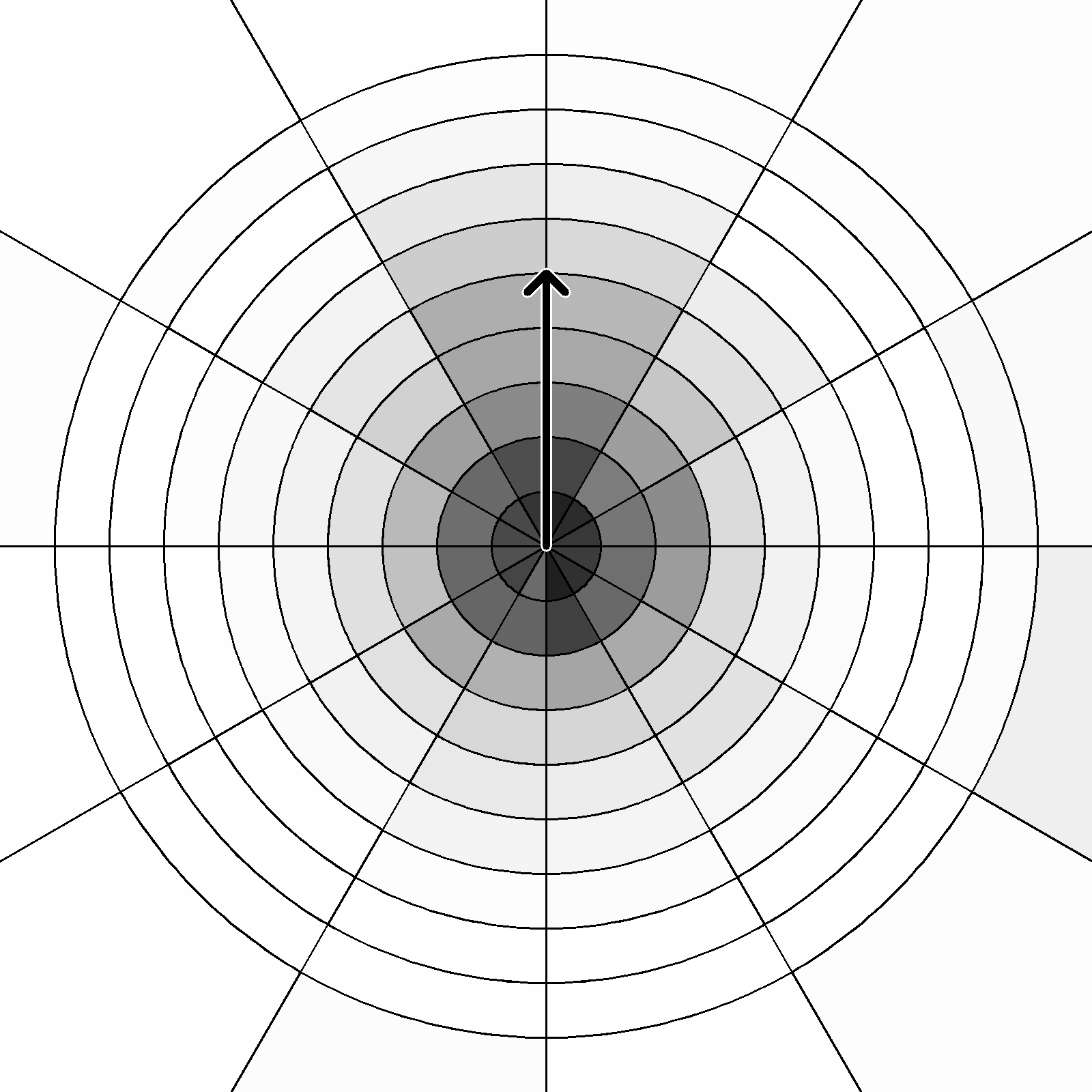}\label{fig:sp_Fd}}
     \myhspace
     \subfloat[][P]{\includegraphics[height=\sizebar]{figures/visibilityPlot/Figure_1.pdf}\label{fig:sp_Fe}}
     \myhspace
     \rulesep
     \myhspace
     \subfloat[][FUL Frontal cone P.\label{fig:sp_Ff}]{\includegraphics[width=\sizescatter]{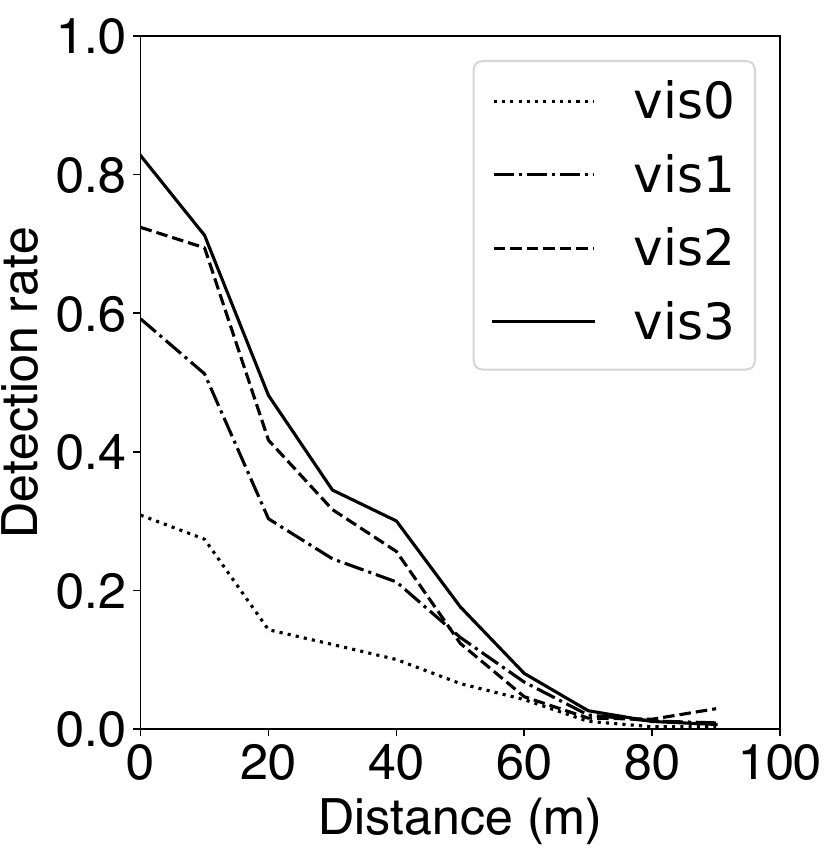}} 
      \caption{Probability P of detection for nuScenes ground truth object processed by Apollo perception module.
      Objects are located around the ego vehicle positioned in the center of the polar grid (radius increments of 10m and angle increments of 30 degrees). The ego vehicle faces upward, as indicated by the arrows. 
      We can observe how P values outline the effective FoV of each sensor setup CAM, LID, and FUL. Moreover, nuScenes occlusion levels have a clear impact on P. As expected, less occluded object (vis3) lead to generally higher detection rates across all grid cells.
      Figures \ref{fig:sp_Cf}, \ref{fig:sp_Lf}, and \ref{fig:sp_Ff} depict the decay of P in the frontal cone for each visibility level based on the sensor setup.}
     \label{fig:polarVisibility}
     
\end{figure*}

\section{Simulation Experiments}

\begin{figure}[t]
\centering
\includegraphics[width=.8\columnwidth]{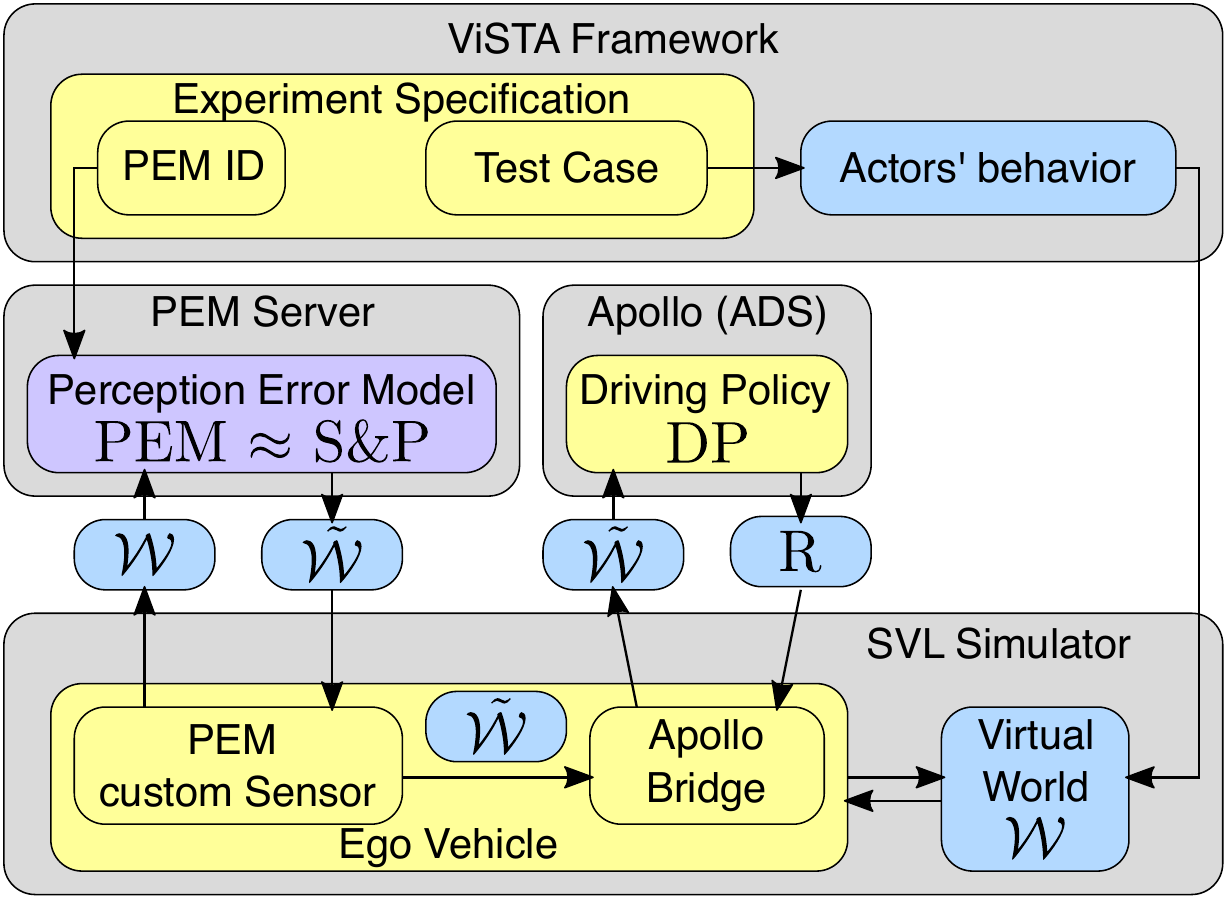}
\caption{Software architecture for virtual testing with PEM integration. The ViSTA framework automates testing by configuring the other components.}\label{fig:swArch}
\end{figure}

\noindent In this section, we present the simulation testing setup aimed at investigating different PEMs, not in raw perception metrics but in terms of their impact on safety.
To this end, we adopted SVL simulator \cite{LGSVL} and Apollo 5.0 \cite{Fan2018} as ADS under test.
Our choice is driven by the open-source nature of the tools, which facilitates our customization, and the reliable co-simulation offered by the SVL bridge towards the CyberRT middleware employed by Apollo. 
We employed the ViSTA framework for testing automation \cite{piazzoni2021vista}. The framework is composed of Python scripts that implement the defined scenarios, control SVL actors in a deterministic manner, and log the results.
We extended ViSTA and SVL simulator by integrating PEMs in the pipeline, as illustrated in \autoref{fig:swArch}.
We prepared a custom sensor in the SVL simulator that connects via socket to a local Python server to send $\W$. The server implements the PEM and processes $\W$ to generate $\OM$ (see  \autoref{eq:pemFunction}) with the determined frequency. This solution provides access to many Python statistical and machine-learning tools.

\subsection{List of Scenario-based Experiments} 
\begin{figure}[t]
     \centering
     \includegraphics[width=0.9\columnwidth]{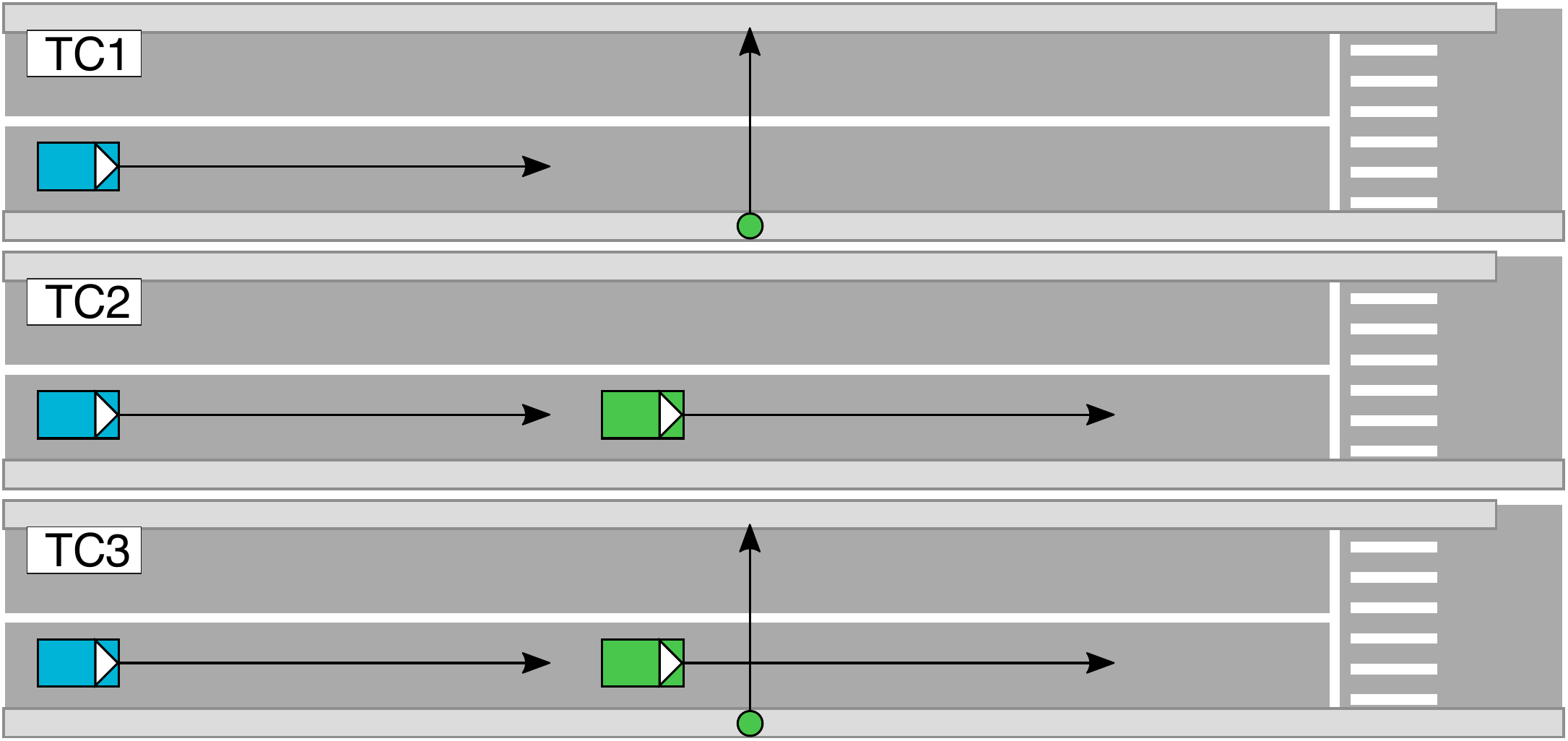}
    \caption{Illustration of the 3 scenarios in our experiments. TC1: jaywalking pedestrian, TC2: following a vehicle, TC3: the combination of both.}
     \label{fig:scenarios}
\end{figure}

We have defined a set of experiments, as illustrated in  \autoref{fig:swArch}.
Each experiment combines a specific Test Case (a Scenario with instantiated parameters) and a PEM.
In this article, we compare the 3 PEMs previously introduced (see \autoref{table:setups}) against 3 representative urban driving scenarios (see \autoref{fig:scenarios}). 

We executed each of the nine resulting experiments 500 times to account for the randomness introduced by the PEMs.
Moreover, we ran each scenario 250 times with error-free perception (baseline), resulting in the ground-truth object list.

\paragraph{Test case TC1} The ego vehicle is driving on a straight road. After ~400m, a pedestrian jaywalks in front of the vehicle. The ego vehicle has to slow down, stop, or perform an evasive maneuver to avoid the collision.
We configured the pedestrian in the simulations so that it causes a collision if the ego vehicle does not react.

\paragraph{Test case TC2} The ego vehicle is driving on a straight road and approaches a leading traffic vehicle driving at 7mps. 
The ego and traffic vehicles proceed for ~500m until reaching the traffic light.
The ego vehicle has the option to overtake the leading vehicle.

\paragraph{Test case TC3} A combination of both previous scenarios. This scenario may also involve occlusion, as the leading vehicle could hide the potential pedestrian detection. While the leading vehicle follows the same script as TC2, the pedestrian starts walking only when this leads to a collision with the ego-vehicle.

These scenarios have been selected as they represent some of the most basic interactions between the ego vehicle and a pedestrian (TC1) or another vehicle (TC2) while involving a potential risk of a collision. Moreover, these scenarios require a similar response to avoid collision.

\begin{figure}[t]
\centering
\includegraphics[width=.9\columnwidth,clip]{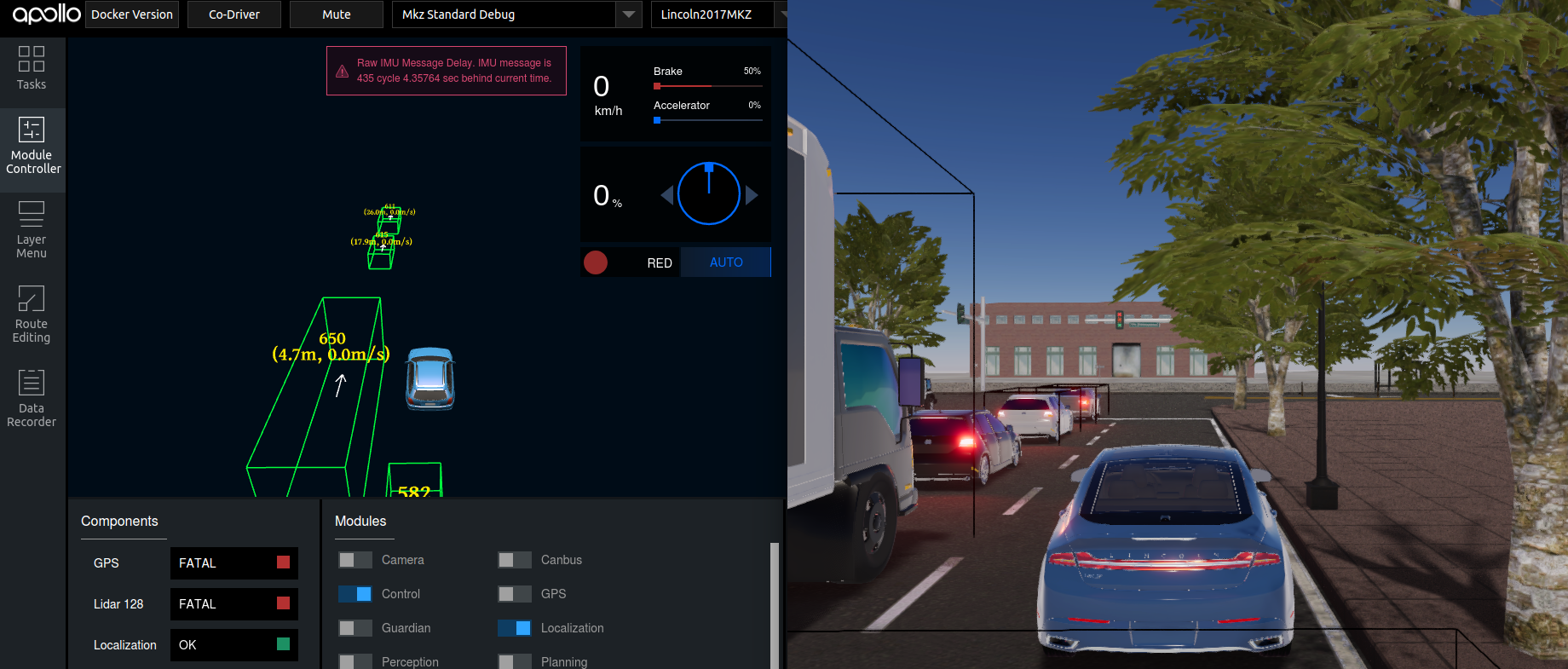}
\caption{Screenshot of the SVL-Apollo co-simulation, including PEM. Bounding boxes in $\OM$ are consistent between Apollo (left) and SVL (right), even undetected objects. 
}\label{fig:cosim}
\end{figure}
\subsection{Experimental Results}
\renewcommand\sizescatter{.15\textwidth}
\renewcommand{\rulesep}{\unskip\ \vrule height 28mm width .01mm}
\renewcommand\myhspace{\hspace{1mm}}
\begin{figure*}[t]
     \centering
     \subfloat[a1][TC1:CAM.]{\includegraphics[width=\sizescatter]{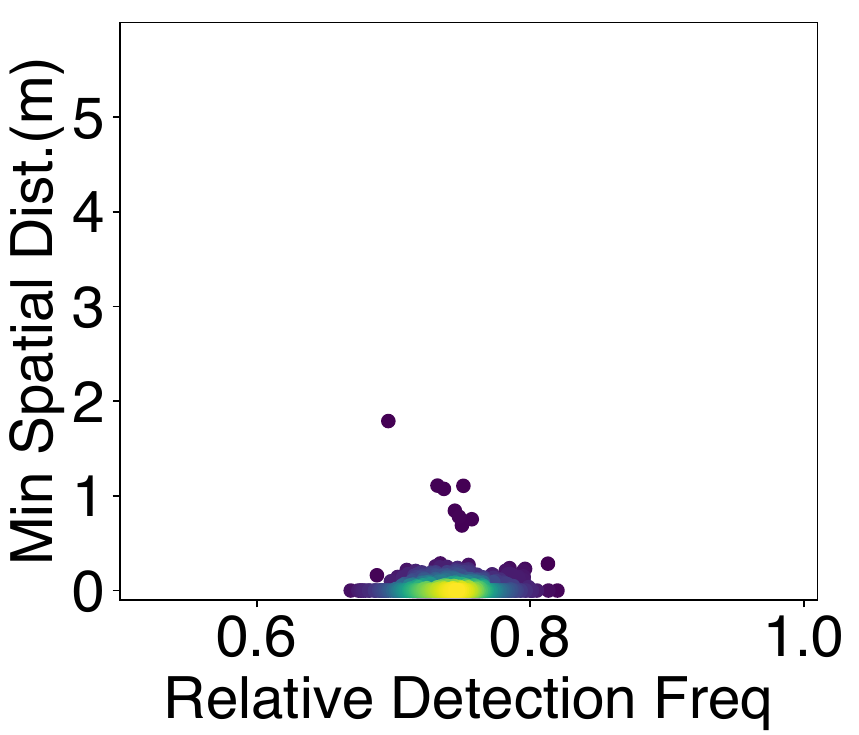}}
      \myhspace
     \subfloat[a2][TC1:CAM.]{\includegraphics[width=\sizescatter]{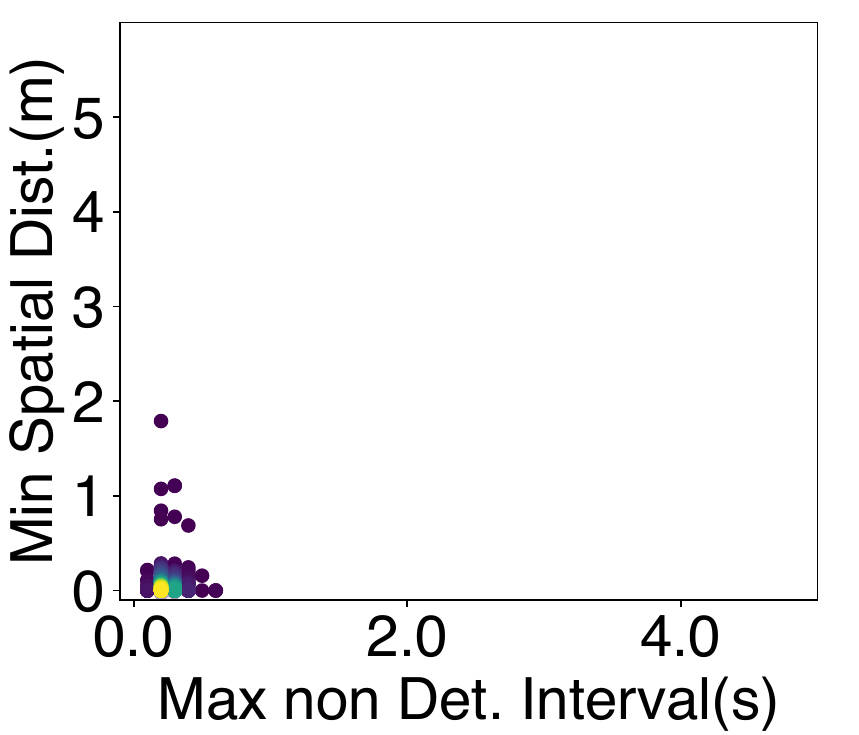}}
      \myhspace\myhspace
      \rulesep
      \myhspace\myhspace
     \subfloat[b1][TC2:CAM.]{\includegraphics[width=\sizescatter]{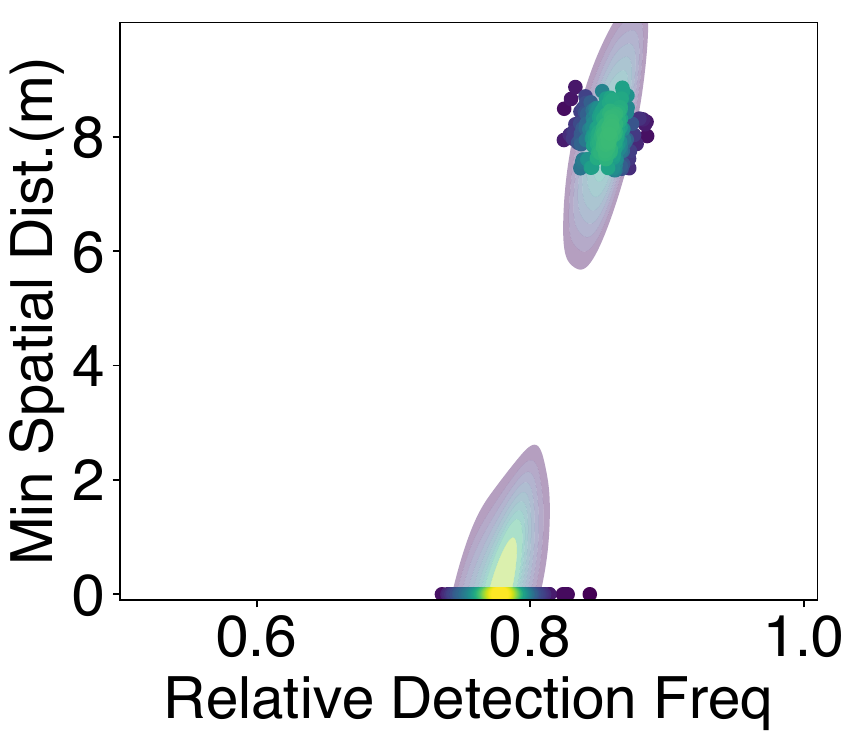}}
      \myhspace
      \subfloat[b2][TC2:CAM.]{\includegraphics[width=\sizescatter]{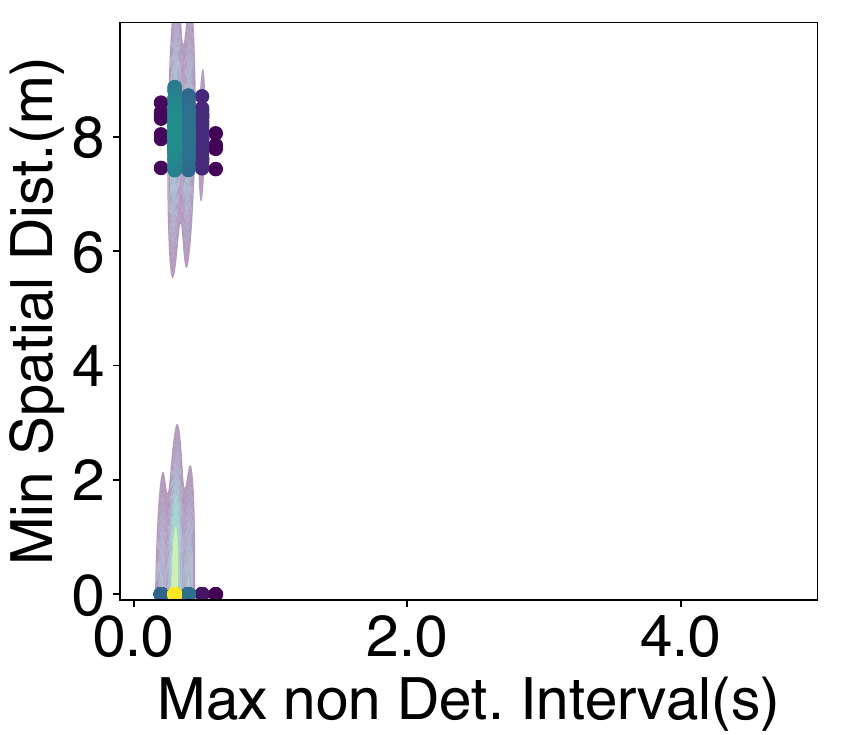}}
      \myhspace\myhspace
      \rulesep
      \myhspace\myhspace
     \subfloat[b1][TC3:CAM.]{\includegraphics[width=\sizescatter]{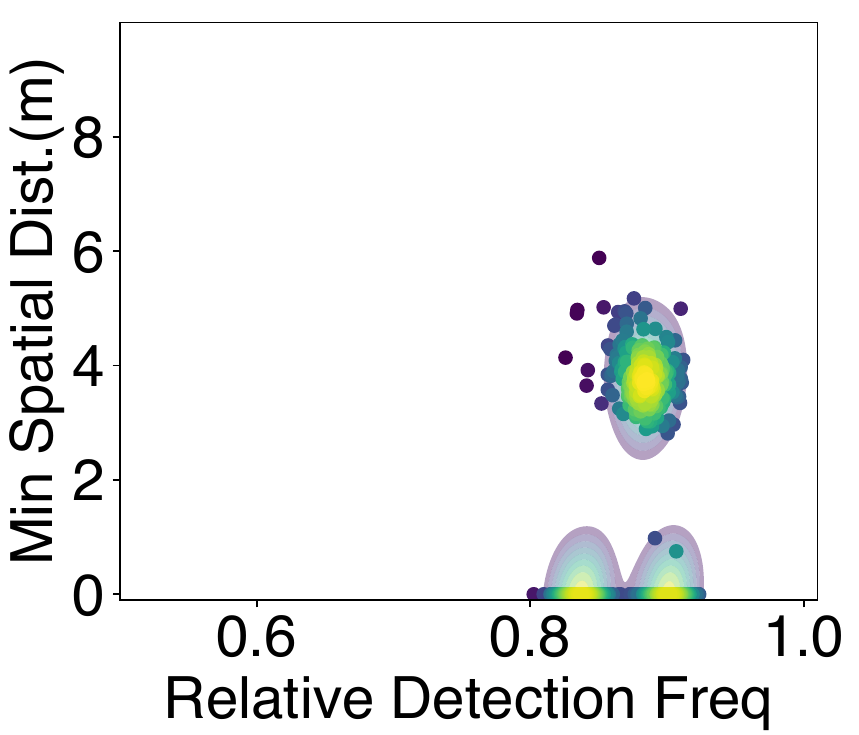}}
      \myhspace
      \subfloat[b2][TC3:CAM.]{\includegraphics[width=\sizescatter]{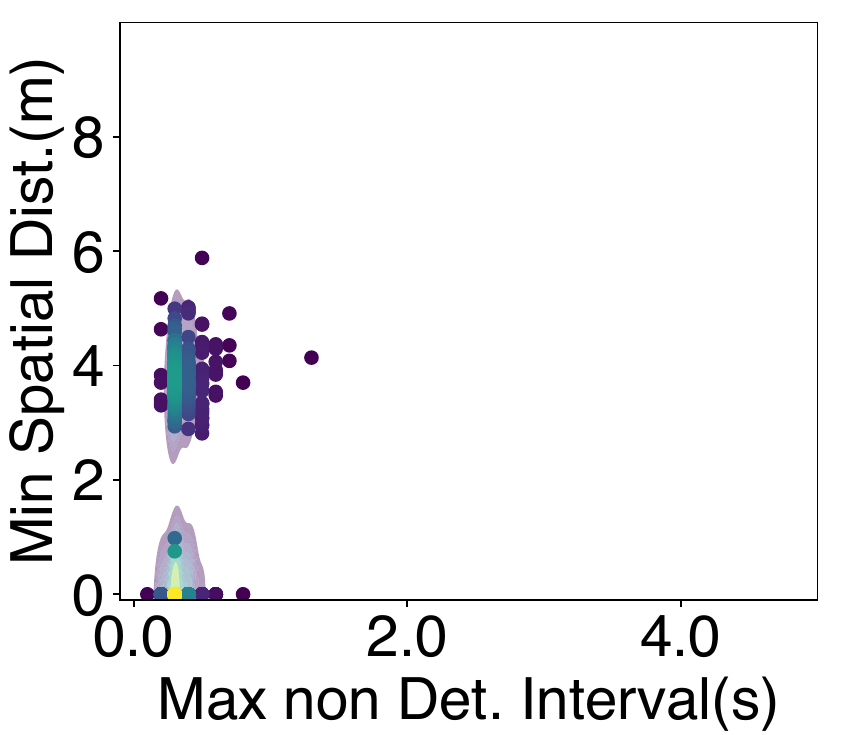}}
      \myhspace 
     \\
     \vspace{-3mm} 
     \subfloat[c1][TC1:LID.]{\includegraphics[width=\sizescatter]{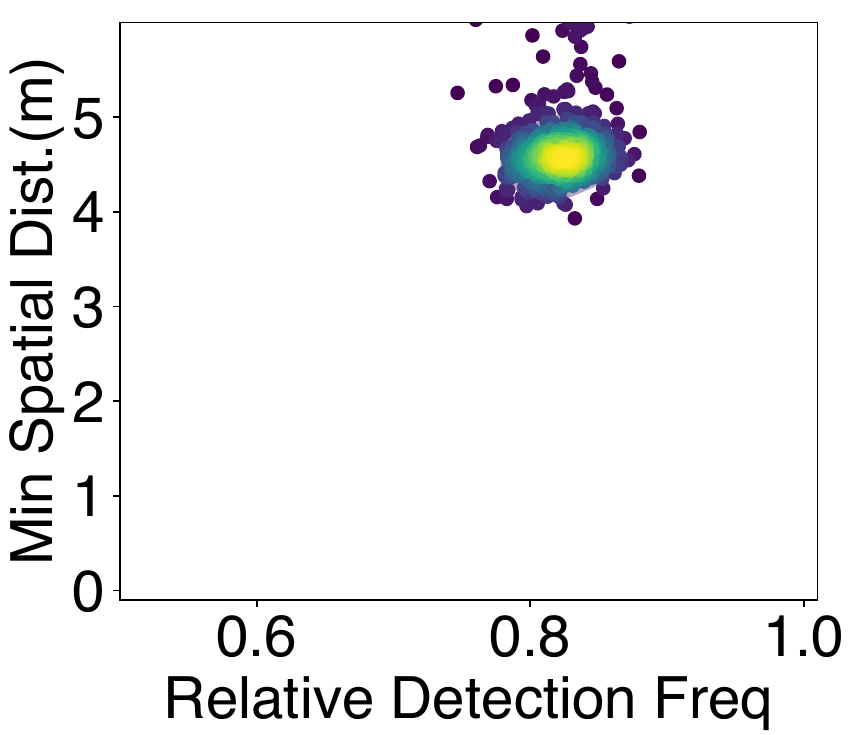}}
      \myhspace
     \subfloat[c2][TC1:LID.]{\includegraphics[width=\sizescatter]{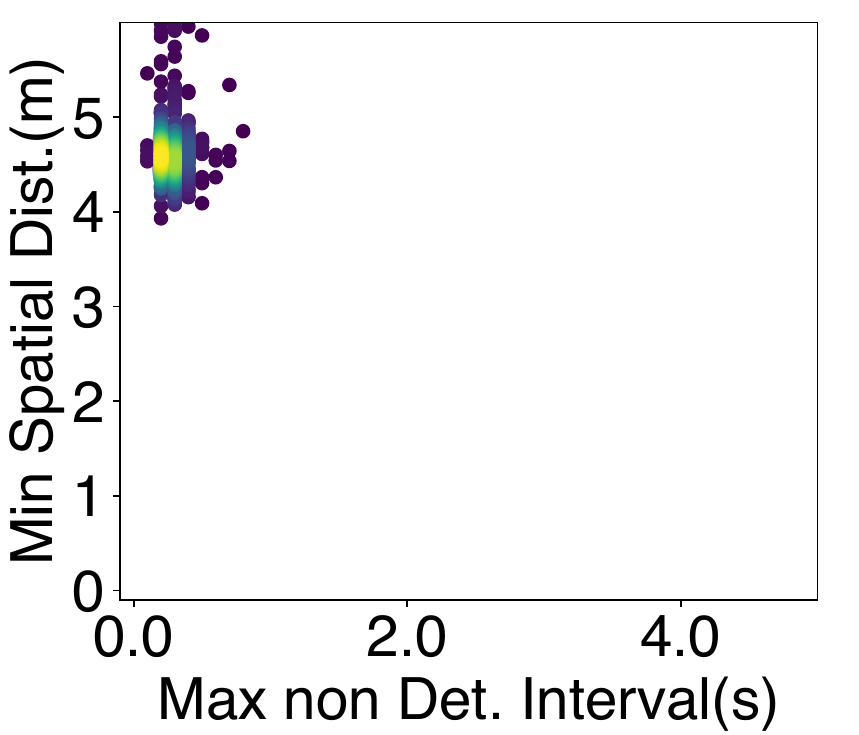}}
      \myhspace\myhspace
     \rulesep
      \myhspace\myhspace
     \subfloat[d1][TC2:LID.]{\includegraphics[width=\sizescatter]{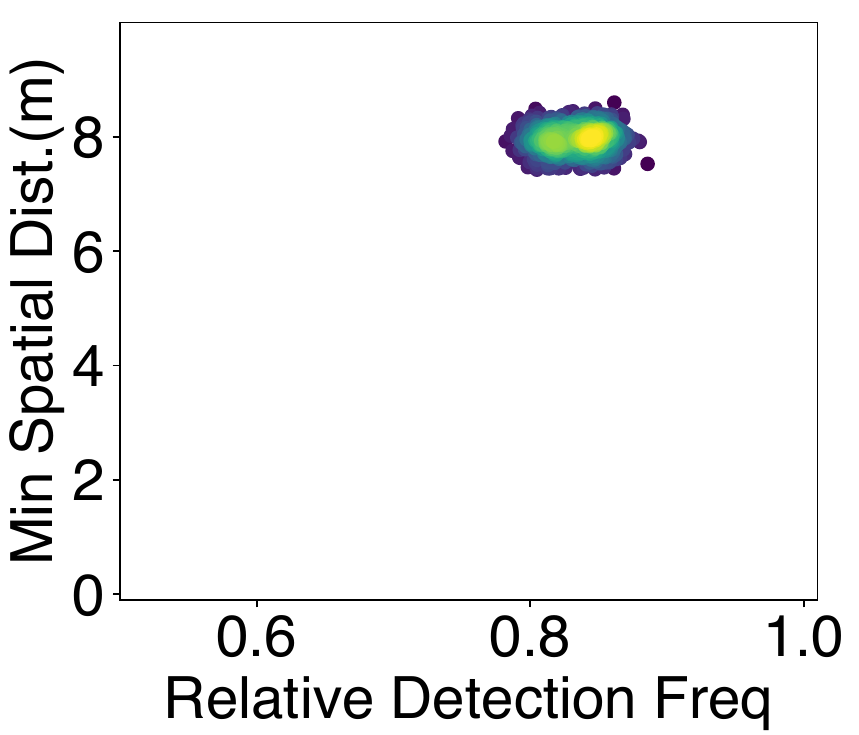}}
     \myhspace
     \subfloat[d2][TC2:LID.]{\includegraphics[width=\sizescatter]{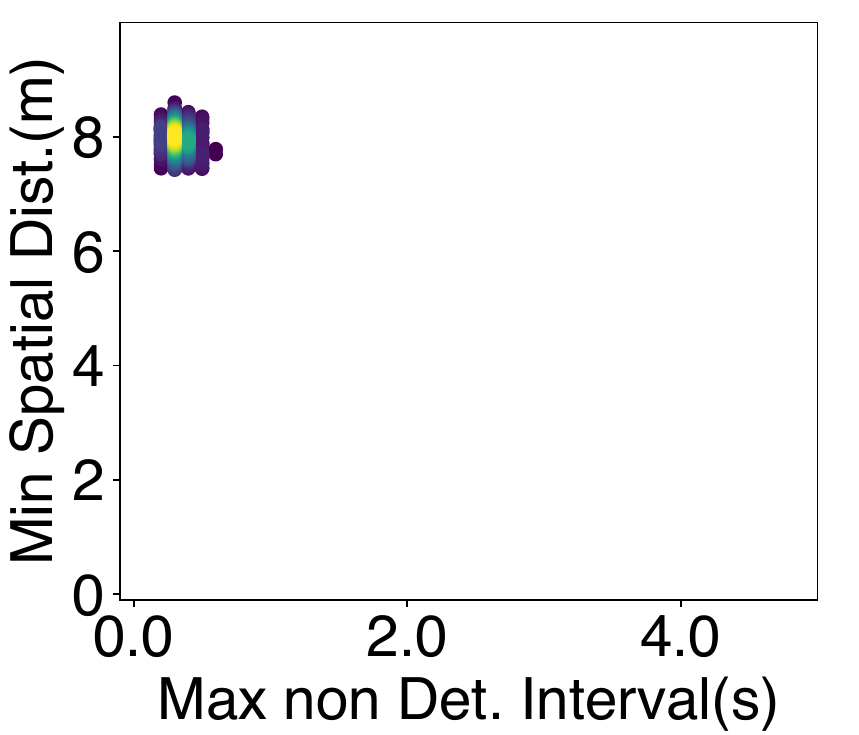}}
     \myhspace\myhspace
     \rulesep
      \myhspace\myhspace
     \subfloat[d1][TC3:LID.]{\includegraphics[width=\sizescatter]{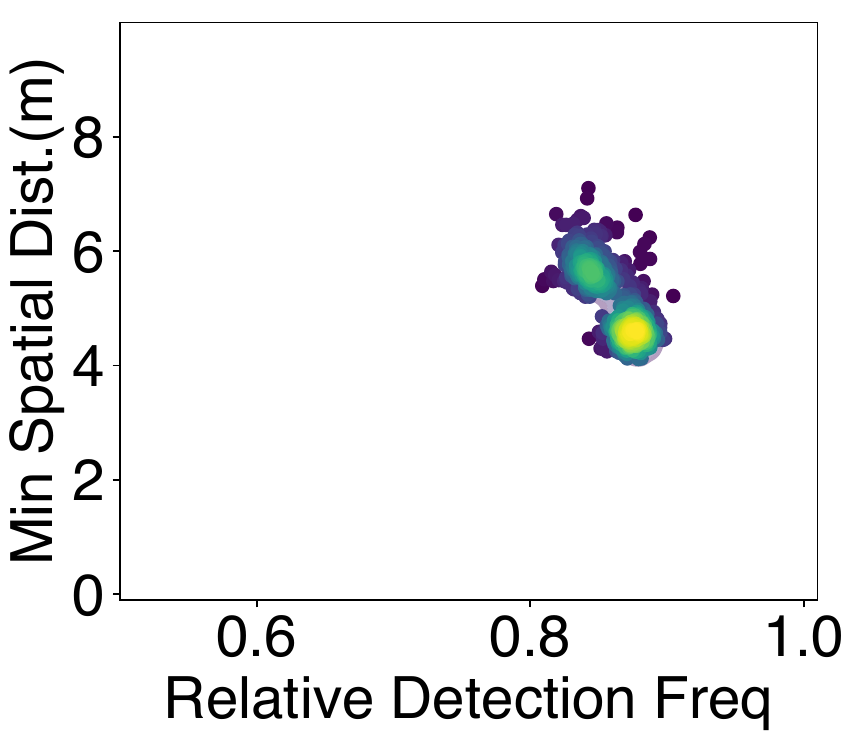}}
     \myhspace
     \subfloat[d2][TC3:LID.]{\includegraphics[width=\sizescatter]{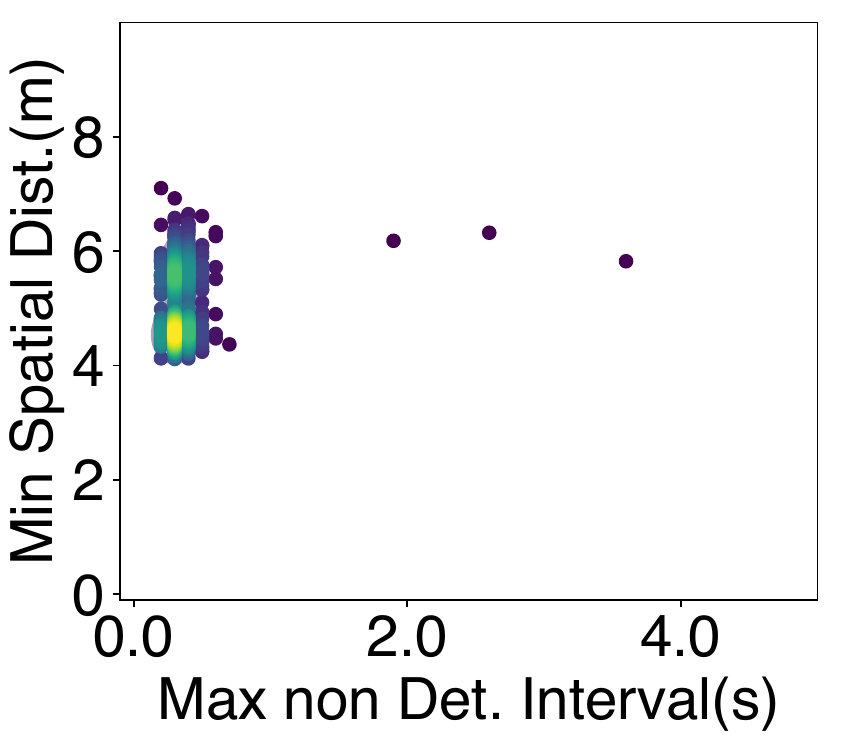}}
     \myhspace
    \\
     \vspace{-3mm} 
     \subfloat[e1][TC1:FUL.]{\includegraphics[width=\sizescatter]{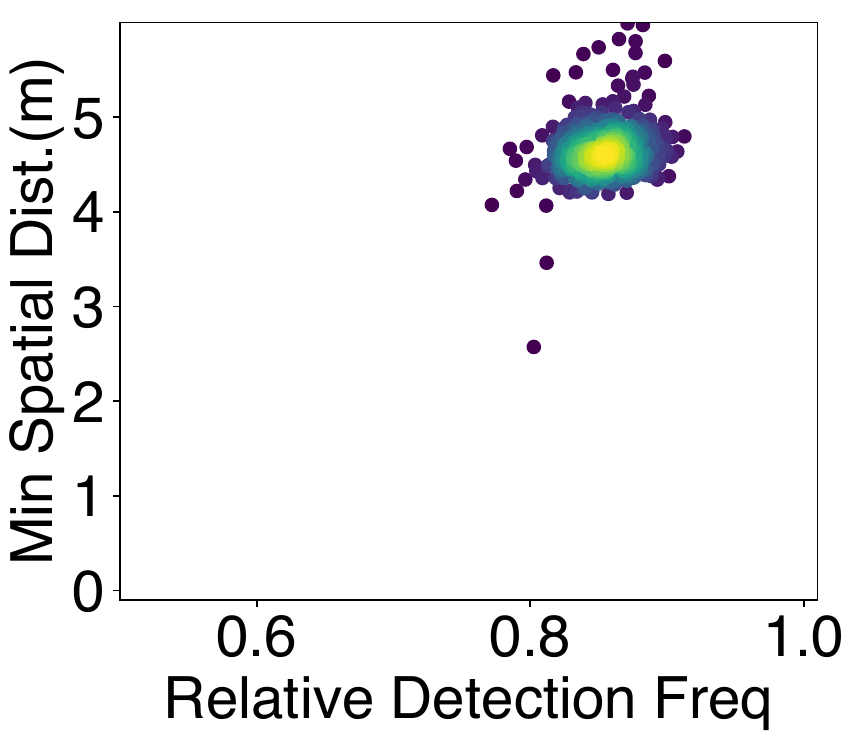}}
      \myhspace
     \subfloat[e2][TC1:FUL.]{\includegraphics[width=\sizescatter]{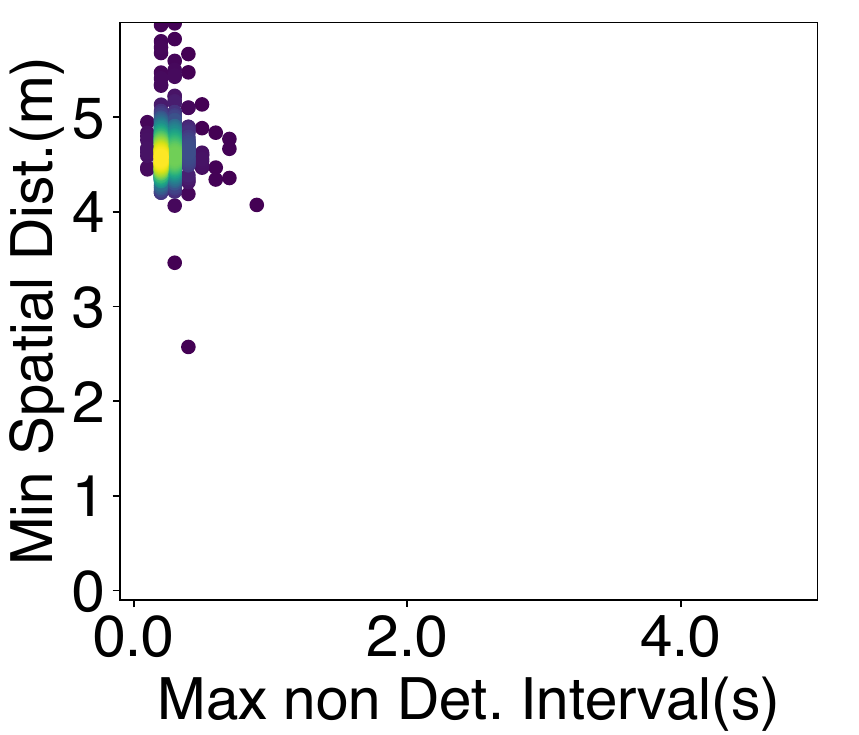}}
      \myhspace\myhspace
     \rulesep
      \myhspace\myhspace
     \subfloat[f1][TC2:FUL.]{\includegraphics[width=\sizescatter]{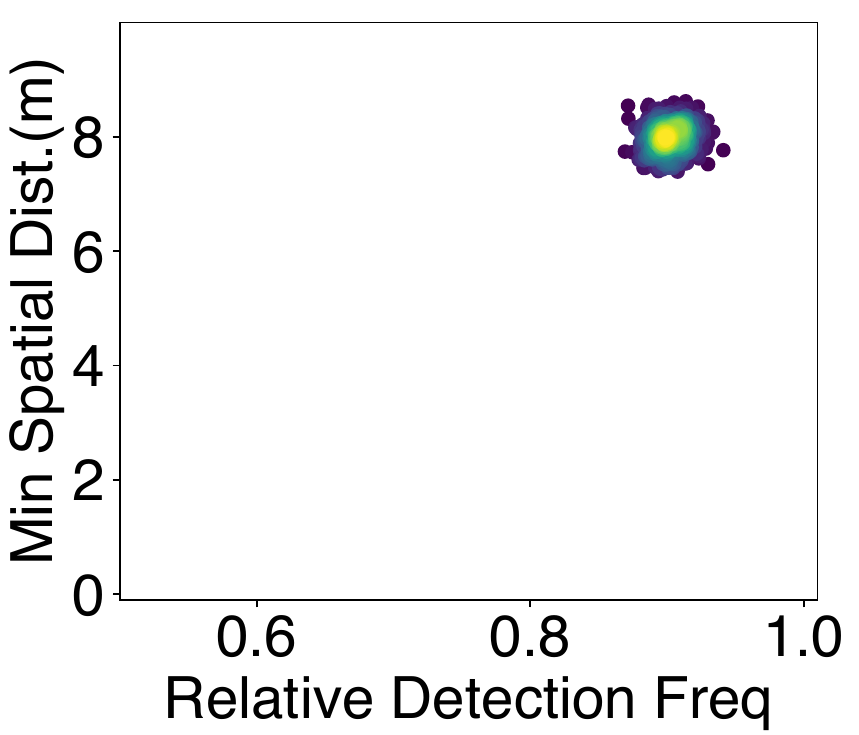}}
     \myhspace
     \subfloat[f2][TC2:FUL.]{\includegraphics[width=\sizescatter]{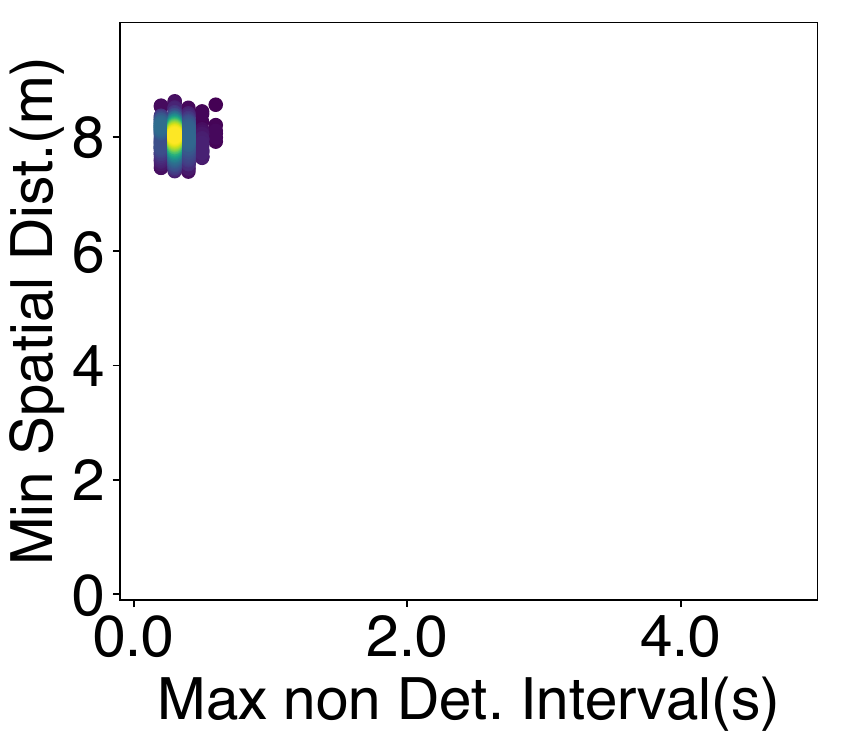}}
     \myhspace
\myhspace
     \rulesep
      \myhspace\myhspace
     \subfloat[f1][TC3:FUL.]{\includegraphics[width=\sizescatter]{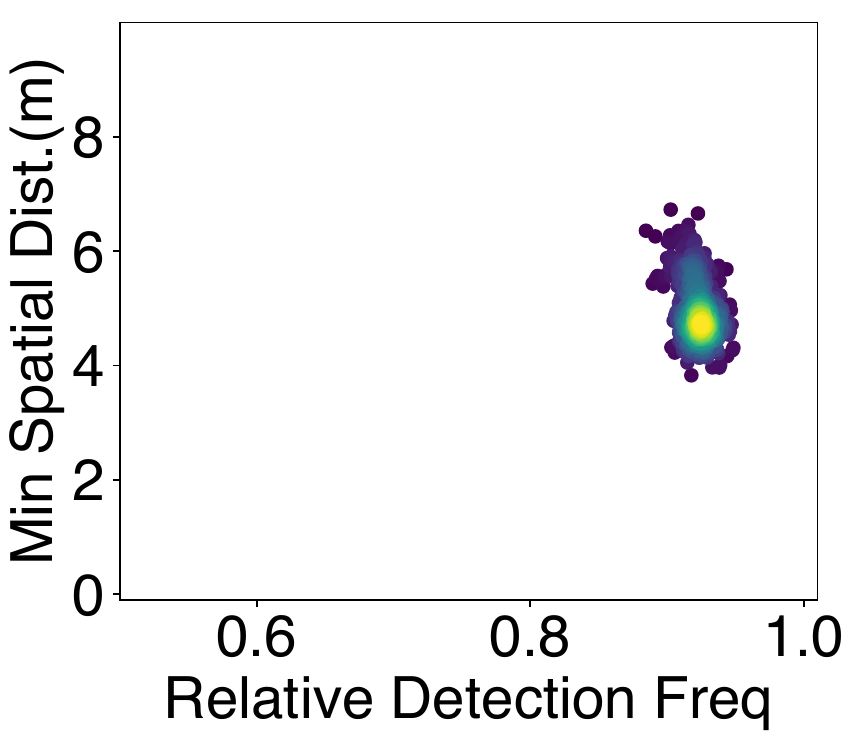}}
     \myhspace
     \subfloat[f2][TC3:FUL.]{\includegraphics[width=\sizescatter]{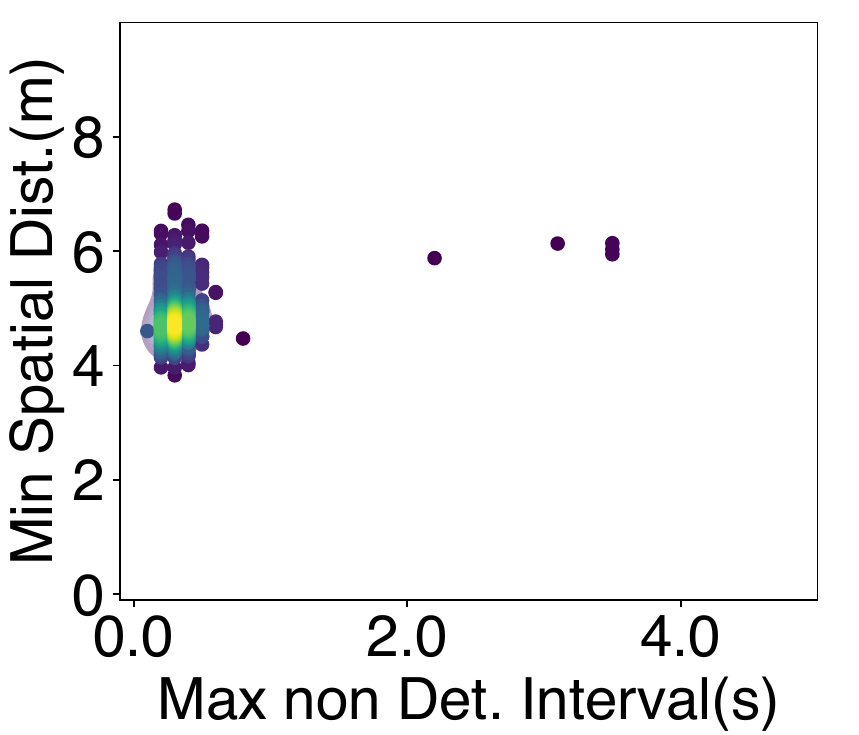}}
     \myhspace
         \\
     \vspace{-3mm} 
     \subfloat[e1][TC1:GT.]{\includegraphics[width=\sizescatter]{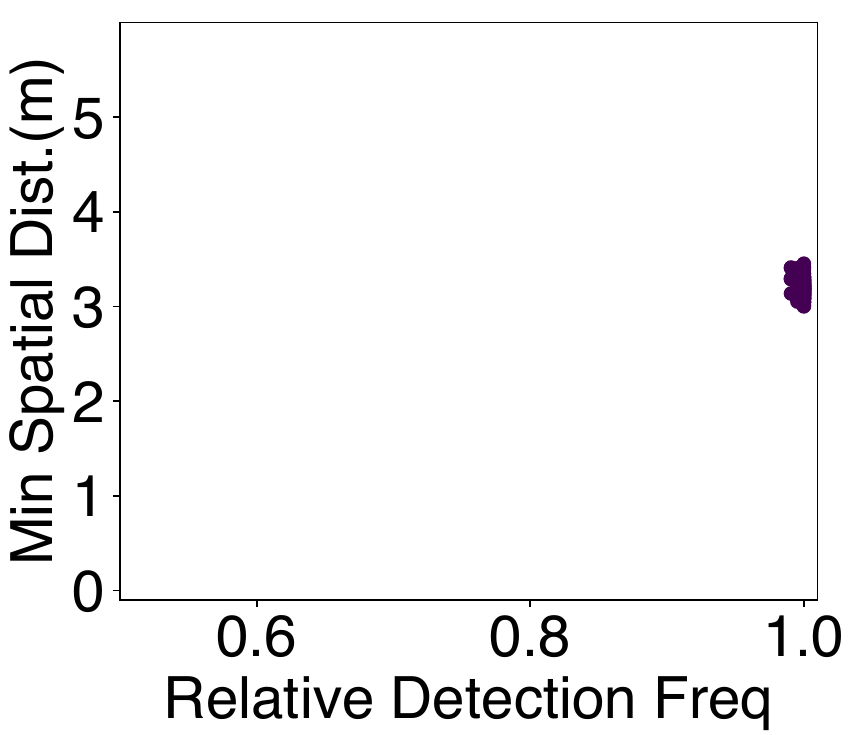}}
      \myhspace
     \subfloat[e2][TC1:GT.]{\includegraphics[width=\sizescatter]{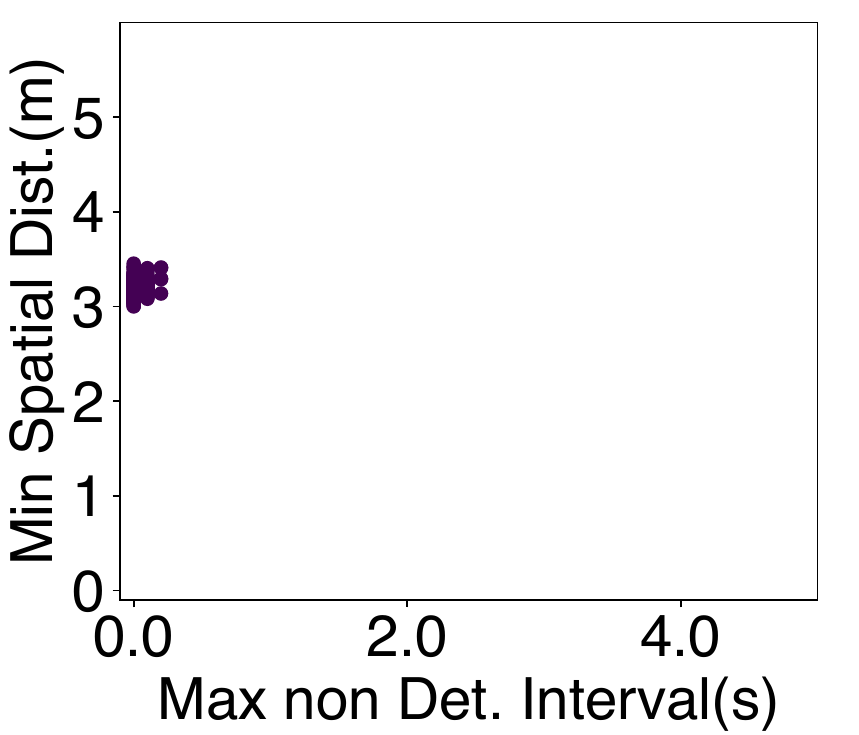}}
      \myhspace\myhspace
     \rulesep
      \myhspace\myhspace
     \subfloat[f1][TC2:GT.]{\includegraphics[width=\sizescatter]{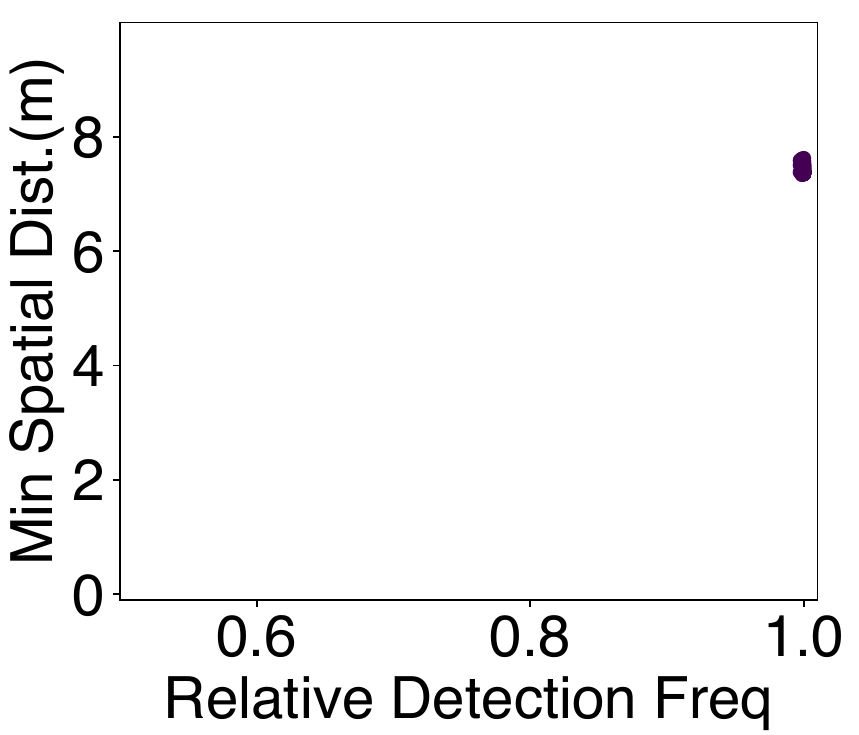}}
     \myhspace
     \subfloat[f2][TC2:GT.]{\includegraphics[width=\sizescatter]{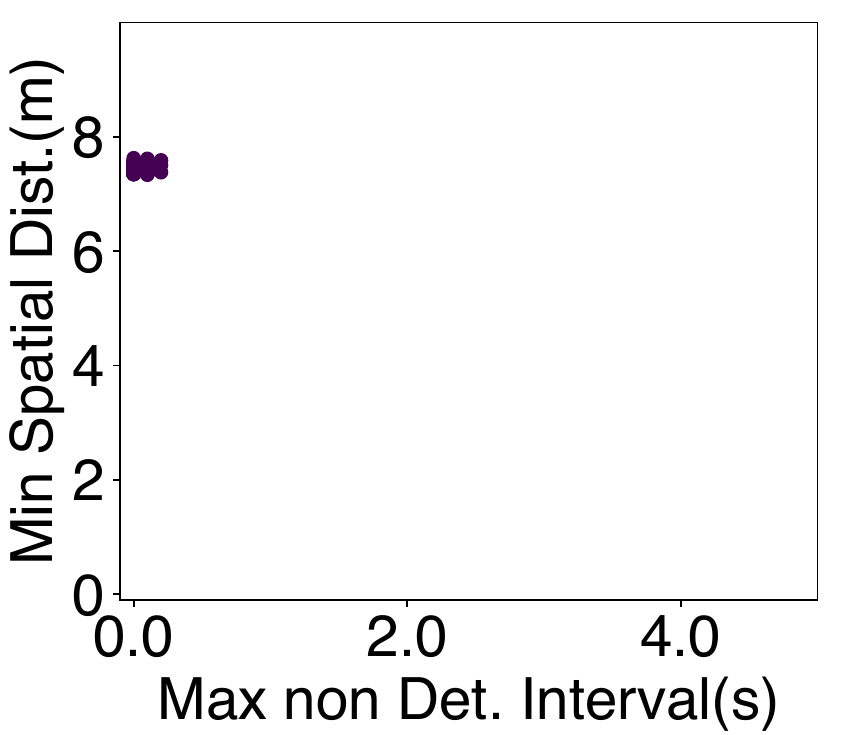}}
     \myhspace
\myhspace
     \rulesep
      \myhspace\myhspace
     \subfloat[f1][TC3:GT.]{\includegraphics[width=\sizescatter]{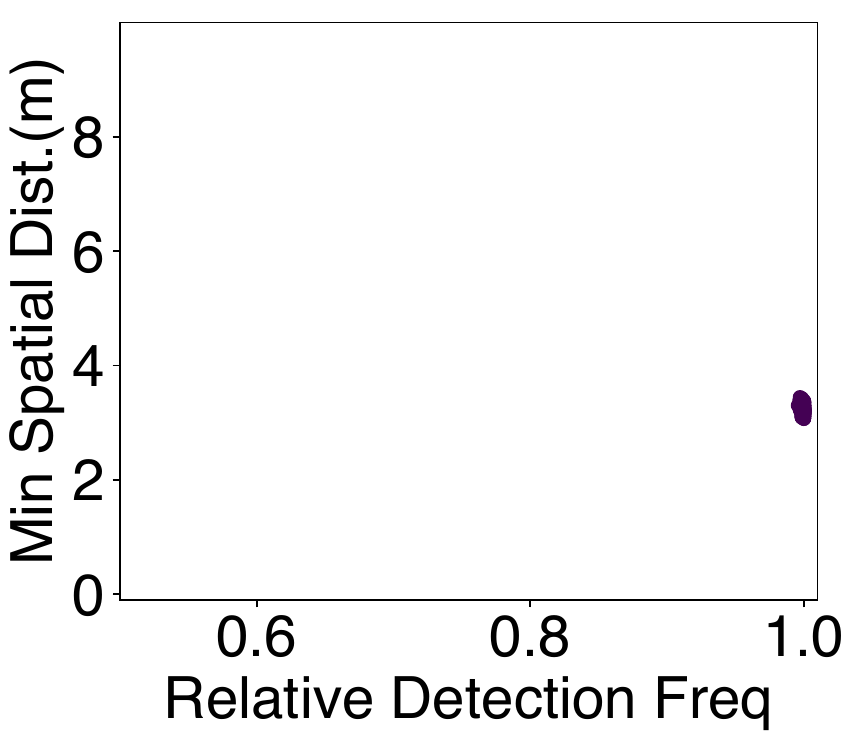}}
     \myhspace
     \subfloat[f2][TC3:GT.]{\includegraphics[width=\sizescatter]{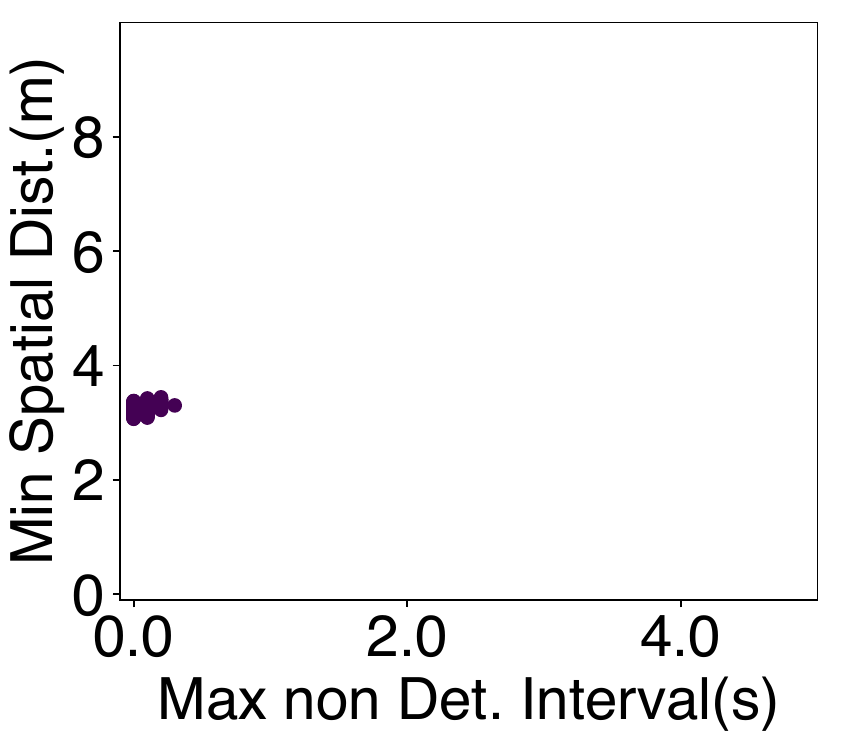}}
     \myhspace
      \caption{Density scatter plots outlining results for each experiment, i.e. a combination of Test Case and PEM.  Columns 1-2: TC1, columns 3-4: TC2, columns 5-6: TC3. Moreover, each row is associated with a particular sensors setup: CAM, LID, and  FUL.
Each couple of plots (e.g., columns 1-2, in the first row) depicts the 500 simulation runs executed under that combination of Test Case and PEM. 
      Density of samples is represented on a color scale from blue (low density) to yellow (high density).}
     \label{fig:scatterplots}
\end{figure*}

In this section, we report the outcome of our experiments, demonstrating the scope of analysis afforded by introducing PEMs in the simulation pipeline.
We compare the effect of deploying the three PEMs shown in \autoref{sec:pemlearnRes}, learned by varying the sensor setup.
Hence, we can observe the safety achieved by each different S\&P.
For ease of understanding, we limit the analysis to one of the most straightforward safety metrics: the minimum distance between the ego vehicle and the obstacle (vehicle or pedestrian) during a complete test case run. Intuitively, we can consider a distance $<1$m as severe as a collision. 
\autoref{table:results} summarize the success rate of each experiment in three different levels of safety based on the minimum spatial distance achieved in each run. 

Additionally, \autoref{fig:scatterplots} depicts the whole set of simulated runs. The vertical axes represent the chosen safety metric (minimum spatial distance) in all plots, while the horizontal refers to a perception metric. 
We highlight two perception metrics, \textit{Relative Detection Frequency}, i.e., the rate of detection of the obstacle, and \textit{Max non-Detection Interval}, i.e., the length of the most extended time interval without a detection. We computed the relative detection frequency of the obstacles only when they were within $100$m of the ego vehicle. This consideration prevents skewed results caused by the amount of time the ego-vehicle spends in the obstacle's proximity compared to the simulation's overall duration.

\begin{table}[t]
\centering
\caption{Success rate by combination of Test Case and Sensor Setup. }
\label{table:results}
 \begin{tabular}{c|r@{\hspace{2ex}}r|r@{\hspace{2ex}}r|r@{\hspace{2ex}}r} 
 Sensor & \multicolumn{2}{c|}{TC1} &
  \multicolumn{2}{c|}{TC2}& \multicolumn{2}{c}{TC3}  \\
 Setup & $<$1m & $>$1m & $<$1m & $>$1m& $<$1m & $>$1m \\ 
 \hline & \\[\dimexpr-\normalbaselineskip+2pt]
 CAM &  99.2\% &  0.8\%&  61.6\% & 38.4\%& 59.0\% & 41.0\% \\ 
 LID &  0.0\% &  100\% &0.0\% &  100\% &0.0\% &  100\%\\
 FUL & 0.0\% &  100\% &0.0\% &  100\% &0.0\% &  100\%\\
\end{tabular}
\end{table}

\subsubsection{TC1}
This scenario is the most challenging, and CAM setup cannot perform safely. Upon further inspection, we noticed a high positional error in the CAM model that prevents the detection of the potential collision, despite the pedestrian being detected at a similar rate compared to LID and FUL.
\subsubsection{TC2}
In this scenario, the AV employing CAM setup presents two distinct behaviors. It can perform safely in $38.4\%$ of the runs. 

\subsubsection{TC3}
Interestingly, the scenario combination of the previous allows for a safer outcome for all setups. In particular, the presence of 2 potential obstacles promotes a slower driving style, leading to lower amount of collisions.

\subsection{Discussion}

We bypass all the considerations presented in \autoref{ssec:criticalities} by employing virtual testing instead of perception metrics to evaluate perception. We can evaluate the overall S\&P by associating it with its safety rather than cumulative metrics with no contextual awareness. Moreover, we are testing a specific Decision Policy implemented by the Decision Module employed in Apollo, which directly addresses I4.
This result is not readily achievable without PEMs, as simulations based on synthetic data are in most cases too computationally expensive and do not provide any direct insight into perception performance (e.g., \autoref{fig:polarVisibility}).

In \autoref{sec:relatedWorks}, we presented some state-of-the-art modeling techniques for perception errors. However, given the relatively under-explored nature of the field, those studies do not follow a standard and structured procedure.
Designing a PEM remains a complex task as it requires formulating various assumptions, such as determining which parameters to consider and which kind of error to model. This makes each study unique and rarely compatible with the others.
However, most of these studies could be easily adapted and integrated into our proposed simulation framework by replacing the PEM in the PEM server illustrated in \autoref{fig:swArch}.  
 
In our previous work \cite{ijcai2020-483}, PEMs were hand-crafted to explore specific \textit{variations}, e.g., varying detection rates.
In this study, PEMs are learned from the nuScenes dataset \cite{Caesar2020} processed by Apollo perception module \cite{Fan2018}. In particular, we selected three sensor setups (CAM, LID, and FUL), subsets of the sensors in the nuScenes dataset.
Moreover, our example PEMs address all four considerations presented in \autoref{ssec:err_model_considerations}. We implement temporal aspects in $M$ and parameter inter-dependencies in $p$. We address positional aspects with the grid partitioning  and occlusion with $\C$, which is not considered in any other related study in the literature.

The main limitation of the proposed approach is the need to define and model each type of error explicitly.
For example, our model does not generate false positives, since errors are only computed on existing objects in the scene. 
Simulations based on synthetic data do not suffer from this problem, since the actual perception module introduces the errors. 
Nevertheless, determining if the false positives generated by the synthetic data are realistic requires modeling error properties such as its distributions and causes. Thus, modeling the errors is also needed for validation purposes.

\section{Conclusion}
 \noindent In this article, we proposed a  generalized data-driven approach to test and study perception errors in a virtual environment.
Our approach involves the introduction of PEM as a computationally efficient way of injecting perception errors into a simulation pipeline.
We provided guidelines for a data-driven modeling for PEM and its integration in a simulation pipeline using open-source software Apollo \cite{Fan2018}, SVL \cite{LGSVL}, and public datasets nuScenes \cite{Caesar2020}.
This approach leads to benefits in many areas. First, it affords a deeper understanding of the perception module, identifying shortcomings and weaknesses. 
Secondly, simulation fidelity increases by not relying on perfect perception, but is not as computationally expensive as relying on synthetic signals.
Additionally, we conducted simulation experiments involving three PEMs, each based on different sensor setups (camera, LiDAR, and a combination of both). Our experiments highlight the connection between S\&P performance and AV safety.
We identify two directions for future developments: PEM fidelity and PEM application. The former involves developing more sophisticated PEMs, considering more object parameters (e.g., size, class) and external conditions (e.g., weather). 
For the latter, we proposed a standardized interface notation for PEM that facilitates comparisons and reusability. These directions can promote the effectiveness of simulation platforms toward a safer deployment of AVs on public roads.

\bibliographystyle{IEEEtran}

\bibliography{IEEEabrv,library}
\end{document}